\documentclass[letterpaper, 10 pt, conference]{ieeeconf}  % Comment this line out if you need a4paper

\IEEEoverridecommandlockouts                              % This command is only needed if 
\usepackage{algpseudocode}

\usepackage{bm}

\usepackage{amsmath}
\usepackage{amssymb}

\title{Multi-gait Locomotion Planning and Tracking for Tendon-actuated Terrestrial Soft Robot (TerreSoRo)}
\author{Arun Niddish Mahendran$^{1}$, Caitlin Freeman$^{1}$, Alexander H. Chang$^{2}$,\\ Michael McDougall$^{3}$, Patricio A. Vela$^{2}$ and Vishesh Vikas$^{1}$% 
\thanks{$1$: The authors are with the Department of Mechanical
Engineering, University of Alabama, Tuscaloosa, AL, USA.
Email: anmahendran@crimson.ua.edu, clfreeman7@crimson.ua.edu, vvikas@ua.edu}%
\thanks{$2$: The authors are with the Institute for Robotics and Intelligent Machines (IRIM) and School of Electrical and Computer
Engineering, Georgia Institute of Technology, Atlanta, GA, USA.
Email: alexander.h.chang@gatech.edu, pvela@ece.gatech.edu}%
\thanks{$3$: The author is with the University of Strathclyde Glasgow, Glasgow, UK.
Email: michael.mcdougall5@gmail.edu}%}
\thanks{This work was supported by NSF grants \#1562911, \#1830432}
}

\usepackage{graphics,graphbox,graphicx} % for pdf, bitmapped graphics files
\usepackage{epsfig} % for postscript graphics files
\usepackage{mathptmx} % assumes new font selection scheme installed
\usepackage{times} % assumes new font selection scheme installed
\usepackage{amsmath} % assumes amsmath package installed
\usepackage{amssymb}  % assumes amsmath package installed
\usepackage{cite}
\usepackage{bm,soul,xcolor}
\usepackage[linesnumbered,lined,vlined,ruled]{algorithm2e}
\usepackage{multirow}

\newcommand{\Fig}{Fig. }
\newcommand{\Sec}{Sec. }

\renewcommand{\Re}{\mathbb{R}}
\newcommand{\MSoRo}{TerreSoRo}
\newcommand{\MSoRoLong}{Tetra-Limb Terrestrial Soft Robot}
\newcommand{\MATLAB}{MATLAB\textsuperscript{\textregistered}}
\newcommand{\Arduino}{Arduino}

\usepackage{algcompatible}
\usepackage{balance}
\newcommand{\textadded}[1]{{#1}}
\graphicspath{./Figures/}

\begin{document}

\maketitle
\thispagestyle{empty}
\pagestyle{empty}

%%%%%%%%%%%%%%%%%%%%%%%%%%%%%%%%%%%%%%%%%%%%%%%%%%%%%%%%%%%%%%%%%%%%%%%%%%%%%%%
\begin{abstract}
The adaptability of soft robots makes them ideal candidates to maneuver through unstructured environments. However, locomotion challenges arise due to complexities in modeling the body mechanics, actuation, and robot-environment dynamics. These factors contribute to the gap between their potential and actual autonomous field deployment. A closed-loop path planning framework for soft robot locomotion is critical to close the real-world realization gap. This paper presents a generic path planning framework applied to \MSoRo~(\MSoRoLong) with pose feedback. It employs a gait-based, lattice trajectory planner to facilitate navigation in the presence of obstacles. The locomotion gaits are synthesized using a data-driven optimization approach that allows for learning from the environment. The trajectory planner employs a greedy breadth-first search strategy to obtain a collision-free trajectory. The synthesized trajectory is a sequence of rotate-then-translate gait pairs. The control architecture integrates high-level and low-level controllers with real-time localization (using an \textadded{overhead} webcam). \MSoRo~successfully navigates environments with obstacles where path re-planning is performed. To best of our knowledge, this is the first instance of real-time, closed-loop path planning of a non-pneumatic soft robot.
\end{abstract}

%%%%%%%%%%%%%%%%%%%%%%%%%%%%%%%%%%%%%%%%%%%%%%%%%%%%%%%%%%%%%%%%%%%%%%%%%%%%%%%%

\section{Introduction}
% 1 - SoRo real-life realization gap
% [VV] Remove 1st sstance
% [VV] Put a short paragraph about the gap and then continue with the challenges
% [VV] - "To best of our knowledge" at beginning or at the end. In the middle, it would get lost.
% [VV] - Merge gaits, and their outcomes as figure* rather than multiple figures
% [VV] - Other gait citations in bibtex
Since the advent of the soft Mckibben actuator, soft materials have been envisioned to be an integral part of next generation robots, including for terrestrial environments. This can be attributed to their ability to adapt and interact with the environment. Over the past few decades there has been active research in discovery of novel soft materials, actuators and sensors \cite{LaEtAl_Science[2016],RusEtAl_Nature[2015]}. Correspondingly, there has been advancement in modeling and control techniques for soft systems like manipulators \cite{ArmEtAl_TRO[2023]}. Additionally, researchers have demonstrated terrestrial locomotion of soft robots using ad-hoc or intuitive gaits \cite{SuEtAl_CurrRobRprts[2021]}. However, in comparison to their rigid counterparts, path planning and navigation of soft robot locomotors is understudied and rarely implemented. This can be primarily attributed to challenges related to modeling of soft materials and their actuation, plus robot-environment interaction. Furthermore, unlike rigid robots, soft terrestrial robots generate lower magnitudes of force when interacting with the environment. This has multiple consequences, including higher variance in locomotive gait displacements (translation and rotation) and higher sensitivity to small environmental changes. 
The research focus is to mitigate the sources of discrepancy between the potential of soft terrestrial robots and their real-life realization by developing a real-time, closed-loop locomotion controller with localization feedback. 

% 2 - Gait Synthesis
 Finite-element modeling and different reduced-order models have been explored by researchers for creating locomotion models \cite{ChangEtAl_RAS[2020],chirikjian1995kinematics,ChangEtAl_IROS[2021],CoEtAl_RoboSoft[2019],BernEtAl_RSS[2019]}. However, the accuracy and predictability of these approaches remains limited.  Models for the friction and sliding of soft materials over a substrate have proven inadequate to capture the robot-environment interaction. Additionally, soft robots are sensitive to manufacturing inaccuracies and defects. Consequently, most locomotion control strategies for soft terrestrial robots rely on biomimetic, intuitive approaches, or trial-and-error \cite{SuEtAl_CurrRobRprts[2021],MaEtAl_JrnofBionicEng[2014]}. More recently, environment-centric, data-driven model-free approaches have been implemented to synthesize gaits \cite{VikasEtAl_IROS[2015],FreemanEtAl_TRO[2023]}. This paper utilizes the gaits synthesized by these approaches as briefly discussed later in the paper.

% 3 - Feedback control and Experimental Setup
Tethered and untethered soft locomotors are actuated using various methods, e.g., pneumatic, shape memory alloys (SMAs), dielectric elastomers (DEA), and motor-tendon actuators\cite{SuEtAl_CurrRobRprts[2021]}. While most use open-loop control strategies, closed-loop control has been performed by few researchers. Patterson et al use a reactive strategy to perform closed-loop control of an SMA-actuated soft swimming robot \cite{PaEtAl_IROS[2020]}; Liu et al \cite{LiuEtAl_IROS[2021]} use a reactive planner for control of a pneumatically actuated soft robot with predetermined gaits; Hamill et al \cite{HaEtAl_ICRA[2019]} perform gait-based path planning using temporal logic; Lu et al \cite{LuEtAl_FrontRobAI[2020]} apply bidirectional A$^*$ with a time varying bounding box. For pose recovery, soft robotics researchers typically rely on economically expensive motion capture systems, e.g., VICON and Optitrack, limiting their more widespread study.
%as it is influenced by factors like marker occlusion. 
This research employs a lattice-based trajectory planner; robot gait models inform the design of of controlled trajectories that move the robot from start to goal while avoiding obstacles. It also details an experimental setup that uses two inexpensive overhead webcams and a localization algorithm that compensates for marker occlusion.

{
Multi-modal motion planning for traditional rigid-body mobile robots typically entails a solution search, through the robot configuration space, for a state sequence (and accompanying control) to accomplish an objective while respecting both task and feasibility constraints. Though articulated mobile robots (e.g. humanoids, multi-legged mechanisms) are often natively described by high-dimensional configuration spaces, planning exploits dimensional-reduction strategies that specialize the search space to modes and configurations relevant to a particular task \cite{DoEtAl_ICRA[2018], BaKaLo_ICRA[2013]_HierApproach_Manip, HaHiGo_ISRR[2011]}; these manifest in conjoinments of several graph-based representations that together capture the multi-modal search space, and where sampling approaches may be utilized to construct graphs or graph-to-graph (i.e. mode-to-mode) transitions. 
Similarly, graph-based representations have been used to synthesize motion plans for vehicles capable of multiple geographically-dependent modalities (e.g. swimming, driving, flying); graphs are synthesized using a sampling-based approach, with edges valued according to modal cost of transport. Dijkstra's algorithm may then be used to identify the optimal multi-modal solution when traveling between locations \cite{SuEtAl_IROS[2020]}.
For a hyper-redundant snake-like robot, a hybrid-optimization approach employs mixed integer programming with model predictive control (MPC) to guide step climbing; the former enforces a particular sequence of discrete modes to be traversed, while the latter designs reference trajectories within each mode \cite{KoTaTa_TCST[2016]}. 
This work focuses on the \MSoRo~soft mobile robot, capable of several locomotion gaits; trajectory planning and re-planning entails a lattice-based search through the space of possible gait sequences, for guiding the robot to desired goal locations within obstacle-strewn environments and in the face of locomotion uncertainty. 
}
% % \hl{System Integration}
% \begin{enumerate}
%     % \item SoRos were supposed to be the next generation robots, even for terrestrial environments. However, this gap has not shrunk. [citation]
%     \item This is "low-force regime" work. [citation]
%     \item Unlike pneumatic, MTAs are low-force generators [citation].
%     \item Challenge specific to SoRos : they rotate with translation, and their localization is often influenced by marker occlusion. The experimental setup compensates for that.
% \end{enumerate}

\textbf{Contributions.} This research, to best of our knowledge, is the first instance of real-time path planning and closed-loop control of a non-pneumatic  soft robot. %The research proposes a generic, closed-loop path planning framework for terrestrial soft robots. 
The experimental setup involves system integration of high-level (online path planning and offline gait synthesis) control, low-level control (\MSoRo~actuation), and real-time pose estimation that uses parallel architecture to process visual feedback. %To the best of our knowledge, this is the first instance of such system integration and experimental validation of closed loop path planning for motor-tendon actuated soft robots. 
Trajectory planning is accomplished using a lattice-based, greedy breadth-first search through the robot's gait control space; motion models of available robot gaits inform the design of controlled locomotion trajectories that move the robot from start to goal while avoiding obstacles. A rotate-then-translate motion control paradigm is adopted to both simplify the procedure and aid tractability of the planning problem. The planned trajectory is re-computed when the position error between the estimated and the actual path exceeds a prescribed threshold.

\textbf{Organization.}
%The paper is structured as follows: 
\Sec \ref{Sec:ProblemFormulation} formulates the navigation problem, details the robot, and provides an overview of data-driven gait discovery and selection. Next, \Sec \ref{Sec:Methodology} discusses the closed-loop control framework, architecture and  path-planning methodology. \Sec \ref{Sec:Experiments} contains the experimental setup, methodology, tracking algorithms, and results. \Sec \ref{Sec:Conclusion} concludes the paper and discusses future work.
%%%%%%%%%%%%%%%

\section{Robot Description and Problem Formulation}
\label{Sec:ProblemFormulation}
\subsection{\MSoRo: \MSoRoLong}

\MSoRo~is a four-limb terrestrial soft robot  actuated using motor-tendon actuators (MTAs). The robot is powered using external power with a low-level controller. The high-level controller and path planner is located off-board on a desktop computer that communicates with the webcam for localization (described in \Sec \ref{Sec:Methodology}). The robot design is the result of topology optimization to allow six identical robots to reconfigure into a sphere\textadded{, whose details are more fully explored in }\cite{FreemanEtAl_SoRo[2021],FreemanEtAl_JCND[2023]_SoRoDesign}. As a result, the limbs are designed for complex geometrical curling and not optimized for any particular locomotion modes. \textadded{The emphasis is on implementing a path planning strategy for individual planar locomotion of the soft robot and does not address reconfigurability.}

\begin{figure}[t]
  \vspace*{5pt}
  \centering
  \includegraphics[width=\columnwidth]{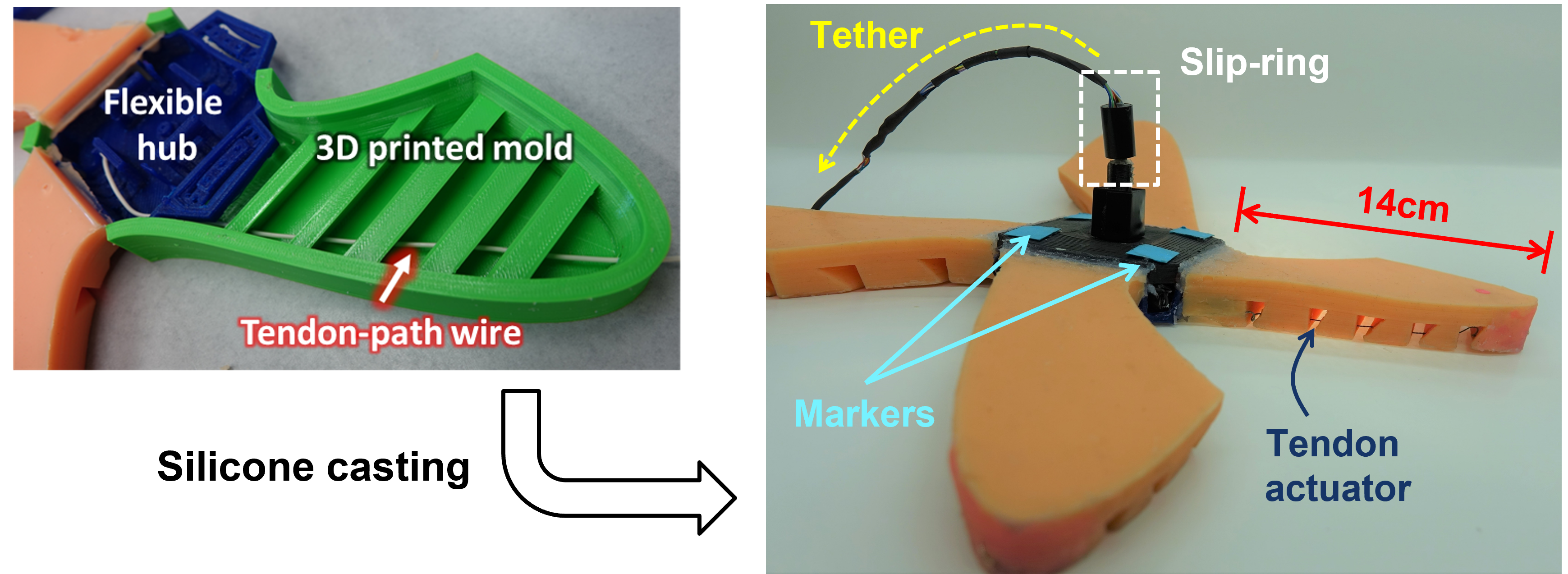}
  \caption{The soft robot is cast using a rigid mold and a tendon-path wire. The motors with spools are placed inside the hub.%
  \label{Fig:robotcasting}}
\end{figure}

Robot fabrication involves integration of soft material limbs, control and actuation payload (motors, electronics), and routing of the tendons through the limb as shown in \Fig \ref{Fig:robotcasting}. The modular fabrication process involves mixing two liquid silicone components (Smooth-On Dragon Skin Part A and B - Shore Hardness 20A) degassed in vacuo. The tendon paths are cast by threading a thick wire through the rigid 3D printed mold as shown in Fig. \ref{Fig:robotcasting} which is removed upon curing of the cast. The central hub is 3D printed with flexible filament (Shore Harness 85A, placed in the mold for casting; the casting is repeated for the other limbs. %
Rapid curling and uncurling of the flexible limbs (\textadded{$450 \text{ ms}$}/transition) is achieved through motor-tendon actuation. Four DC motors with 3D printed PLA spools are placed in the hub and secured using zip ties. Teflon tubing is inserted into the individual fins of each limb to prevent tear caused by the difference in stiffness between the silicone and the fishing line tendon, \Fig \ref{Fig:robotcasting}. Finally, threaded fishing line attached to the spool is routed through each fin and anchored at the end with a fishing hook. A slip ring is incorporated into the tether connector to reduce any effects of built-up torsion in the tether. %Design parameters of the fins of each limb (separation, height, thickness, etc.) are experimentally determined to maximize curling while still permitting uncurling upon relaxing the motor.

% However, all possible motion primitives are quasistatic and all robot configurations are statically stable, satisfying the assumptions of the MFC framework. Hence, this robot can be used to highlight and validate this gait synthesis procedure as (1) simple actuation produces complex movement with coupled translation and rotation, (2) the stick-slip nature of the limb movement complicates traditional gait synthesis and increases the dependence of the robot behavior on the environment, and (3) the exploration of a diverse set of gaits can be used for future planned open-loop path planning. %
%%%%%%%%%%%%%%%%%%%%%%%%%%%%%%%%%%%%
\subsection{Gait Synthesis}
\label{SubSec:GaitSynthesis}
The gaits for \MSoRo~are synthesized (discovered) using an  environment-centric framework that discretizes the factors dominating the robot-environment interaction. The procedure of synthesizing locomotion gaits is described briefly.\footnote{The reader may refer to \cite{VikasEtAl_IROS[2015],FreemanEtAl_TRO[2023]} for detailed analysis, and to  \cite{bollobas_modern_2013} for the graph theory terminology and concepts used.}

\begin{figure}[h]
    \centering
    \includegraphics[width=\columnwidth]{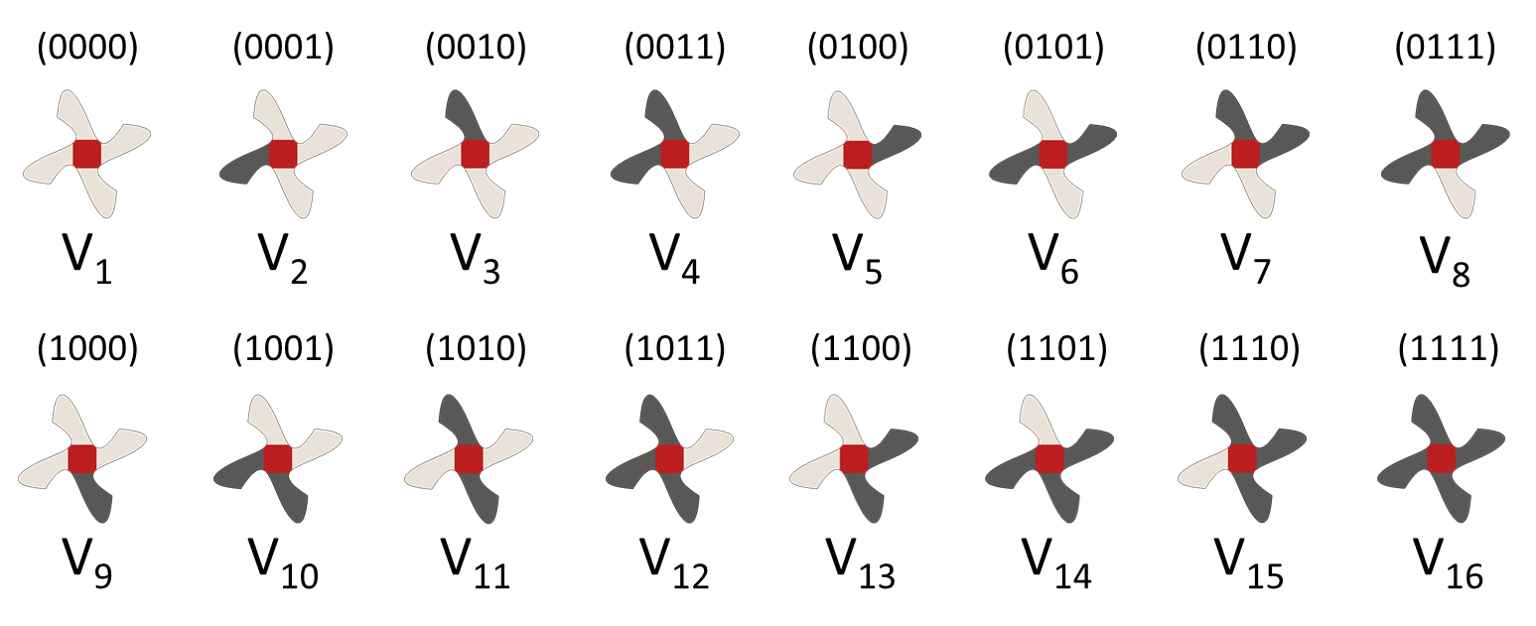}
    \caption{The robot states represented by a four digit binary number. Each digit corresponds to one of the four limbs to indicate if the limb is actuated ($1$/ black color limb) or un-actuated ($0$/ light color limb).}
    \label{Fig:RobotStates}
\end{figure}

\smallskip \noindent{\bf Environment-centric Framework.} Conceptually, locomotion results from optimization of forces acting at different parts of the body that ultimately effect change in inertia \cite{Radhakrishnan_PNAS[1998]}. In that context, \textit{robot states} are defined as discrete physical states (e.g., postures, shapes) where the forces acting on the robot body in each state are considerably different. % Concisely, they discretize the factors dominating the robot-environment interaction. (this is repeated a few sentences earlier)
\textit{Motion primitives} refer to the possible transitions between these robot states. These transitions result in motion of the robot (rotation and translation). A weighted digraph is effective in modeling the robot states, motion primitives, resulting motion, and their inter-dependencies. The robot states and motion primitives correspond to the digraph's $n$ vertices $V(G)$ and the $m$ directional edges $E(G)$, respectively. For \MSoRo, robot states correspond to permutations of the four limbs being curled (actuated) or uncurled (un-actuated) as illustrated in \Fig \ref{Fig:RobotStates}. For this robot, actuation is binary (on/ off) and all possible permutations (states) are statically stable. The weight associated with each edge $e_i$ is the resulting motion of the motion primitive, i.e., the translation $\bm{p}_i\in\Re^{2\times 1}$ and rotation $\theta_i$ measured in the coordinate system of the initial vertex of the edge. For this discussion, the edge weight $\bm{w}_i$ is modeled as a normal distribution with mean $\bm{\mu}_{i} \in \Re^{3\times 1}$ and covariance matrix ${\Sigma}_{i} \in \Re^{3 \times 3}$:

\begin{align} 
% \bm{w}_{i} = \vGaussian{\mu_{i}}{\Sigma_{i}} \label{Eqn:ArcWeight}
\bm{w}_i = \mathcal{N}\left(\bm{\mu}_i,\Sigma_i\right)
,  %
\bm{\mu}_i(e_i) = \begin{bmatrix}\bm{p}_i\\ \theta_i
\end{bmatrix},  
{\Sigma}_i(e_i) = \begin{bmatrix}
{\Sigma}_{pp} &\Sigma_{p\theta}\\
\Sigma_{\theta p} & \Sigma_{\theta\theta}
\end{bmatrix}.
\label{Eqn:ArcWeight}
\end{align}

In summary, we define the following matrices with elements corresponding to edges $e_i$.
\begin{center}
\begin{tabular}{|c|c|c|}
\hline
    & Description & Dim \\ \hline \hline
     $P(E)$& mean displacement matrix, $P(e_i)=\bm{p}_i$ & $\Re^{2\times m}$\\\hline
     $\Theta(E)$& mean rotation matrix, $\Theta(e_i)=\theta_i$ & $\Re^{1\times m}$\\\hline
    \multirow{2}{*}{$S_p(E)$}& translation covariance trace matrix & \multirow{2}{*}{$\Re^{2\times m}$}\\
    & where $\displaystyle S_p(e_i) = \mathrm{tr}\left(\Sigma_{pp}(e_i)\right)$ &\\ \hline
    \multirow{2}{*}{$S_\theta(E)$} & rotation covariance matrix & \multirow{2}{*}{$\Re^{1\times m}$} \\ & where $S_\theta(e_i) = \Sigma_{\theta \theta}(e_i)$&\\  \hline
\end{tabular}
\end{center}

\textit{Learning of the environment} is equivalent to learning the graph edge weights. Experimentally, this is achieved by traversing all the edges of the graph without repeating any and recording the resulting motion. This traversal sequence, referred to as the Euler cycle, is repeated multiple (five) times with randomized starting positions and path orders to learn the probabilistic weights of the graph.

\smallskip \noindent{\bf Translation and Rotation Gaits.}  %A closed walk consists of a sequence of vertices starting and ending at the same vertex, where consecutive vertices in the sequence are connected by a directed edge. A simple cycle is a closed walk where no vertices or directed edges are repeated, other than the start and the end vertex. 
Locomotion gaits are defined here as simple cycles that are transformation invariant. The transformation invariance principle implies that the distance and rotation of the robot are preserved irrespective of starting vertex as it traverses through all edges of the simple cycle. It has been proven that under this definition, there will exist two types of planar gaits: translation and rotation \cite{FreemanEtAl_TRO[2023]}. The former is the gait when the cumulative rotation of the simple cycle is zero. The latter corresponds to the simple cycle when translation of all the edges is zero. Hence, we individually optimize for these two type of gaits with different cost functions and constraints.

The \textit{gait library} comprises the synthesized translation and rotation gaits discovered using the discussed data-driven approach (learning of the graph and searching for optimal gaits). The cost functions, $J_t,J_\theta$ linearly weight the locomotion, variance and  gait length while assuming small rotations of the motion primitives.
\begin{equation}
\begin{gathered}
    J_t(\mathbf{z}) = % \bm{\alpha}_t^T\underbrace{P\mathbf{z}}_{\mathrm{translation}} + \beta_t \underbrace{S_{p}^T\mathbf{z}}_{\mathrm{variance}} + \gamma_t \underbrace{{1}_{1\times m}\mathbf{z}}_{\mathrm{length}}\\
    \left(\bm{\alpha}_t^T P + \beta_t S_{p} + \gamma_t 1_{1\times m} \right)\mathbf{z}\\
    J_\theta(\mathbf{z}) = \left({\alpha}_\theta \Theta + \beta_\theta S_{\theta} + \gamma_\theta 1_{1\times m} \right)\mathbf{z}
\end{gathered}
\label{Eqn:costfuncs}
\end{equation}
where $\{\bm{\alpha}_t,\beta_t,\gamma_t,\alpha_\theta,\beta_\theta,\gamma_\theta\}$ are the linear weights, and the binary vector $\mathbf{z}\in \{0,1\}^m$ is the mathematical representation of a gait. Consequently, the gait synthesis is formulated as a \underline{B}inary \underline{I}nteger \underline{L}inear Programming (BILP) optimization problem with linear constraints that can be solved using optimization solvers, e.g., MATLAB\textsuperscript{\textregistered}. The translation $\mathbf{z}_t$ and rotation $\mathbf{z}_\theta$ gaits are synthesized using
\begin{align}
\begin{gathered}
    \mathbf{z}_t = \min_{\mathbf{z}} J_t(\mathbf{z}) \quad \mathrm{s.t.} \quad\Theta \mathbf{z}\leq \varepsilon_\Theta\\
    \mathbf{z}_\theta = \min_{\mathbf{z}} J_\theta(\mathbf{z}) \mathrm{~s.t.~} |P\mathbf{z}| \leq \varepsilon_p \forall z_i=1\\
    \mathrm{Gait~constraints:} B\mathbf{z}=0, B^i\mathbf{z}\leq 1, z_i\in\{0,1\} \forall i,\\
    \nexists \mathbf{z}_1, \mathbf{z}_2 \mathrm{~s.t.~} \mathbf{z} = \mathbf{z}_1+\mathbf{z}_2,~B\mathbf{z}_1=B\mathbf{z}_2=0
\end{gathered}
\label{Eq:constraints}
\end{align}
where $B$ is the incidence matrix, $B^i$ is the positive elements of $B$, and the gait constraints mathematically ensure that vector $\mathbf{z}$ is a simple cycle. 

Once the gait library has been created and contains a synthesized rotation gait and translation gait, symmetry of the robot can be assumed to expand the library to improve path planning capabilities. The permutations of the translation gait w.r.t. each limb are considered for control purposes to achieve change in orientation of 90 degrees. Such behavior is observed in biology (e.g., brittle star) \cite{As_JrnlofExpBio[2012]} where the animal can change their leading limb to change the direction of their translation. While it is assumed that these permutations will have similar motion in four different directions, each of these gaits are tested to characterize their distinct twists.
\begin{figure*}[h]
\vspace*{0.065in}
{\scriptsize \textbf{(a)}}\hspace*{-0.2in}     \includegraphics[width=0.6\textwidth,align=t]{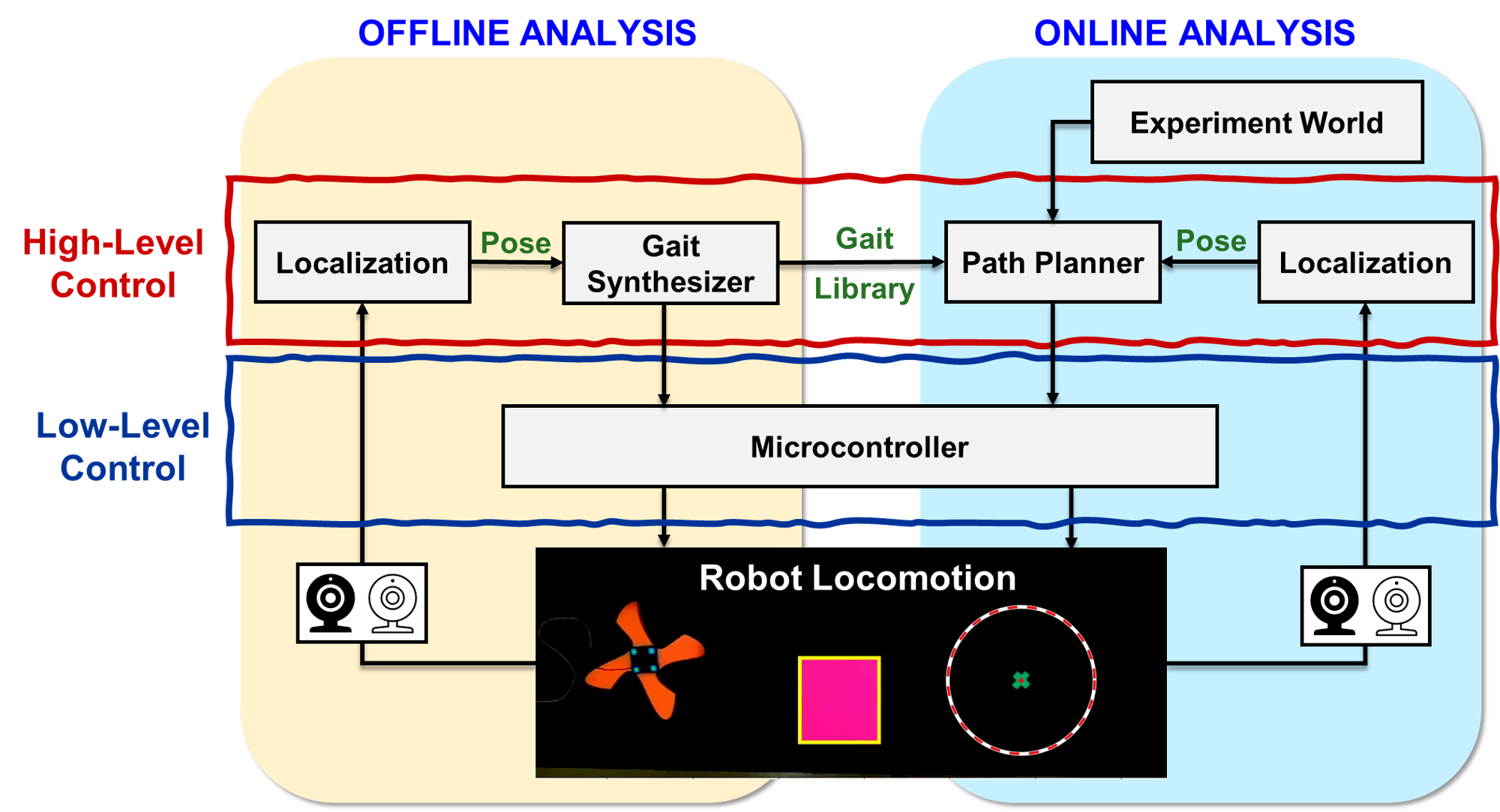}     {\scriptsize \textbf{(b)}} \hspace*{0.1in}\includegraphics[height=2.5in,align=t]{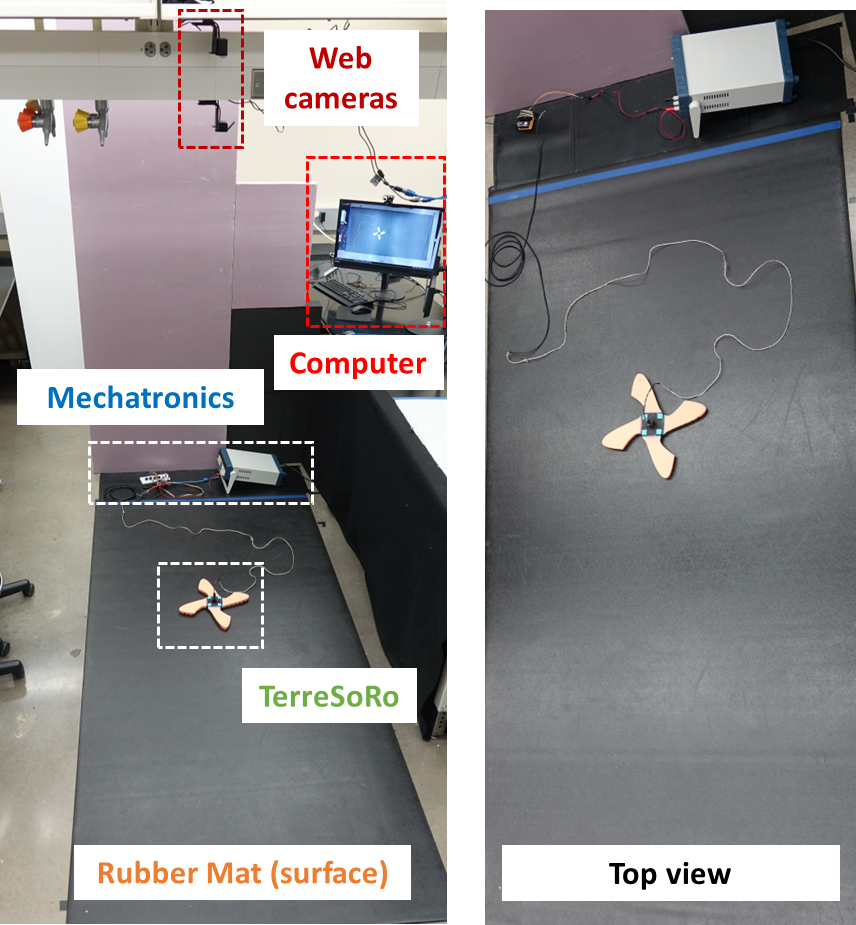}
    \caption{(a) The control architecture combines offline analysis to generate the Gait Library for the online path planning with localization feedback. (b) The experimental setup involves system integration of high and low-level controllers involving processing in MATLAB and microcontroller.}
    \label{Fig:ControlArchitecture}
\end{figure*}
%%%%%%%%%%%%%%%%%%%%%%%%%%%%%%%%%%%%
\subsection{Problem Statement}
%Way-point navigation between two points with obstacle-avoidance with multiple (finite) gait library, where each gait has a probabilistic twist (mean and variance). 
% \hl{Soft mobile robots, such as \MSoRo, manipulate their deformable body composition to accomplish distinctly useful locomotion, relative to more traditional rigid-body robots. In particular, locomotion results from unique interactions between their soft body material composition and the surrounding environment. This allowance comes with distinct challenges; locomotion outcomes are heavily coupled to manufacturing and material variabilities, factors that often are difficult to control. Data-driven techniques are demonstrably effective for: (1) discovering useful gaits indigenous to a particular manufacturing instance, and (2) generate motion models that characterize these gaits for planning and control. Progression of these mobile robots toward autonomous field deployment entails an ability to both plan viable trajectories through an environment as well as accomplish some form of feedback-based control to track synthesized plans.  }

% \hl{We employ a data-driven approach to synthesize gaits and predictive motion models for \MSoRo; these inform a lattice-based planner that designs gait sequences, and accompanying locomotion trajectories, that move \MSoRo~ to prescribed goal locations within an obstacle-strewn environment. Feedback takes the form of trajectory re-planning, when tracking error exceeds a pre-defined threshold.}

Soft mobile robots, such as \MSoRo, manipulate their deformable body to accomplish distinctly useful locomotion, relative to more traditional rigid-body robots. In particular, locomotion results from unique interactions between their soft body material composition and the surrounding environment. This allowance comes with distinct challenges; locomotion outcomes are heavily coupled to manufacturing and material variabilities, factors that often are difficult to control. Data-driven techniques are demonstrably effective for: (1) discovering useful gaits indigenous to a particular manufacturing instance, and (2) generating motion models that characterize these gaits for planning and control. Progression of these mobile robots toward autonomous field deployment entails an ability to both plan viable trajectories through an environment as well as accomplish some form of feedback-based control to track synthesized plans.

We employ a data-driven approach to synthesize gaits and predictive motion models for \MSoRo; these inform a lattice-based planner whose gait sequence outputs, and accompanying locomotion trajectories, move \MSoRo~to prescribed goal locations within an obstacle-strewn environment. Feedback takes the form of trajectory re-planning when tracking error exceeds a pre-defined threshold.

%%%%%%%%%%%%%%%

\section{Methodology, Framework and Algorithm}
\label{Sec:Methodology}

The control framework architecture comprises a high-level controller in \MATLAB~that communicates with the low-level \Arduino~microprocessor to result in locomotion of the robot as summarized in \Fig \ref{Fig:ControlArchitecture}. Localization is achieved by analyzing the webcam output and tracking the center of the robot. In the offline state, the Gait Synthesizer uses the motion data from the Euler cycle experiments to build the Gait Library. The gaits therein are then experimentally validated to store the expected motion data to feed into the path planner. In the online state, the experiment world (which maps the obstacles and initial robot pose) and the gait library are used to initialize the path planner. Real-time control is then achieved by comparing the expected pose from the path planner and the instantaneous experiment pose to inform the low-level controller and re-planning is performed as necessary.
\subsection{Localization}
\label{Subsec:Localization}
%The experimental setup consists of the the \MSoRo~on a rubber garage mat, and two overhead webcams. Low-level robot control is performed by the \Arduino and is integrated with high-level planning and localization that is performed in MATLAB\textregistered. The first webcam is used to capture HD video of the experiments. The second webcam has its properties (e.g., contrast, brightness, etc.) adjusted to allow for better image segmentation of the four neon markers on the robot hub to facilitate tracking. Similarly, the obstacles, and goal are identified. The real-time robot pose is estimated as described in \Sec \ref{Subsec:Localization}. The Gait Library described in \Sec \ref{Subsec:GaitSynthesizer} informs path planning described in Sec. \ref{Subsec:TrajectoryPlanning}.

The feedback to the robot controller plays a critical role in path re-planning. As can be seen in the experimental setup, \Fig \ref{Fig:ControlArchitecture}b, two overhead webcams are used. One webcam is used to record HD video; the other is used for localization and has its properties (e.g., contrast, brightness, etc.) adjusted to facilitate image segmentation of the four neon markers on the robot hub. Both webcam videos are processed in parallel using the \MATLAB~Parallel Processing Toolbox\textsuperscript{\texttrademark}.
%The experiment performs localization of the robot using two web-cameras that are processed in parallel using the \MATLAB Parallel Processing Toolbox. The web-cameras have specific tasks - one is dedicated for localization purposes, while the other records HD video. Each of the video stream is processed by separate cores on the microprocessor. 
The ``localization core" video is processed through a image mask that highlights the markers located on the robot (blue markers on orange robot), obstacles (pink) and the target (green cross), as seen in \Fig \ref{Fig:ControlArchitecture}. This video is stored and a pollable data queue accesses the data for localization at appropriate times. The pose estimation is performed using Arun's method \cite{arun1987least}, which finds the least-squares solution to the pose by using singular value decomposition. Occlusion of the markers by the tether is managed by identifying the marker in the next time frame using the nearest neighbor. The occluded markers are reconstructed using the estimated pose. 
\textadded{Camera capture occurs at $\sim 30$Hz. 
The robot pose estimation provides feedback at $6$Hz on an Intel\textsuperscript{\textregistered} Xeon\textsuperscript{\textregistered} E5-1650 v4, as dictated by the computational complexity and the robot's average motion profile.}
%every 5th frame out of the 30 frames per second is been processed for real time robot’s pose estimation.}

% \begin{algorithm}[h]
% \caption{Robot pose estimation}\label{Alg:PoseEstimation}
% \KwData{Two sets of markers $\bm{p}_i,\bm{p}_i'$ related by\\$\displaystyle \bm{p}_i'=Rp_i+\bm{t}+\bm{n}_i,~\forall i=1,2,\cdots, N$}
% \KwResult{$R \in SO(3), ~\bm{t}\in\Re^{3\times 1}$}
% { $\displaystyle \bm{p} \gets {\frac{1}{N}\sum_{i=1}^N \bm{p}_i}, \quad \bm{p}' \gets {\frac{1}{N}\sum_{i=1}^N\bm{p}_i'}$\;
% $\displaystyle \bm{q}_i \gets {\left(\bm{p}_i-\bm{p}\right)}, \quad  \bm{q}_i' \gets {\left(\bm{p}_i'-\bm{p}'\right)}$\;
% $\displaystyle H \gets {\frac{1}{N}\sum_{i=1}^N \bm{q}_i \bm{q}_i'}$. 
% Find SVD of $H =U\Lambda V^T$\; 
% $R \gets {VU^T}, \quad \bm{t} \gets {\bm{p}'-R \bm{p}}$\;
% }
% \end{algorithm}
\begin{figure*}[h]
\vspace*{0.065in}
    % \centering
    \textbf{\scriptsize(a)} \hspace*{0.24\textwidth}\textbf{\scriptsize(b)}\hspace*{0.34\textwidth}\textbf{\scriptsize(c)}\\
    %\vspace*{-5pt}
    \includegraphics[width=0.25\textwidth, align=c]{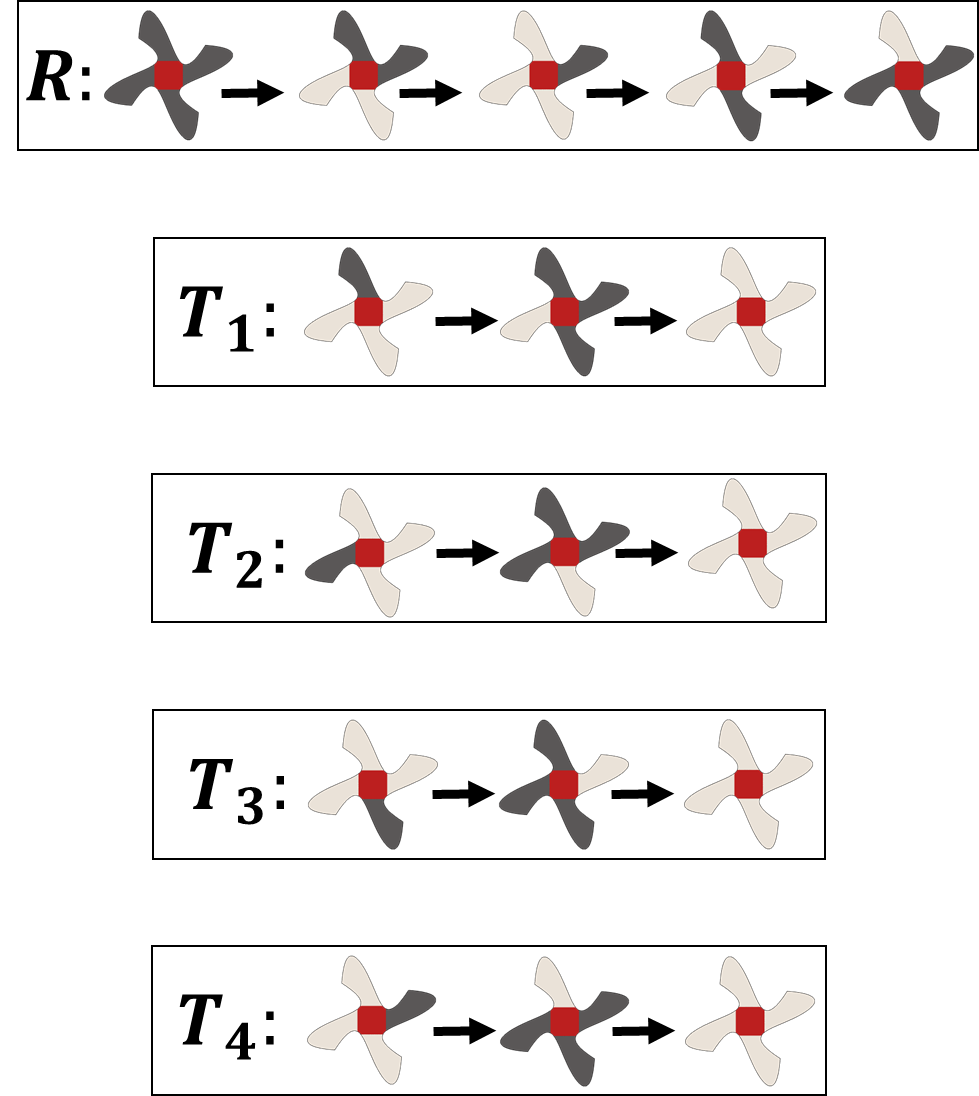}
    \includegraphics[width=0.35\textwidth, align=c]{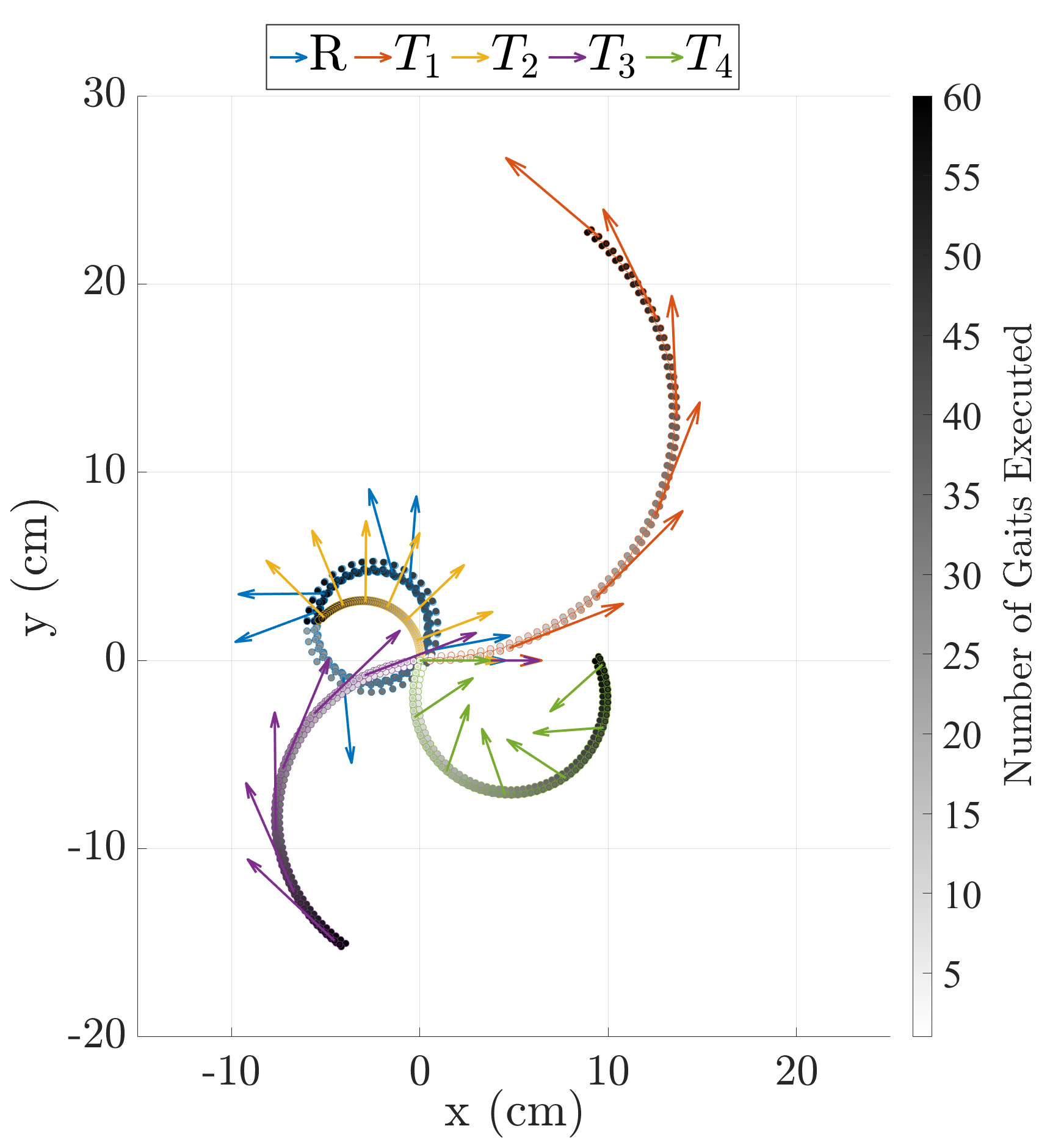}
    \includegraphics[width=0.35\textwidth, align=c]{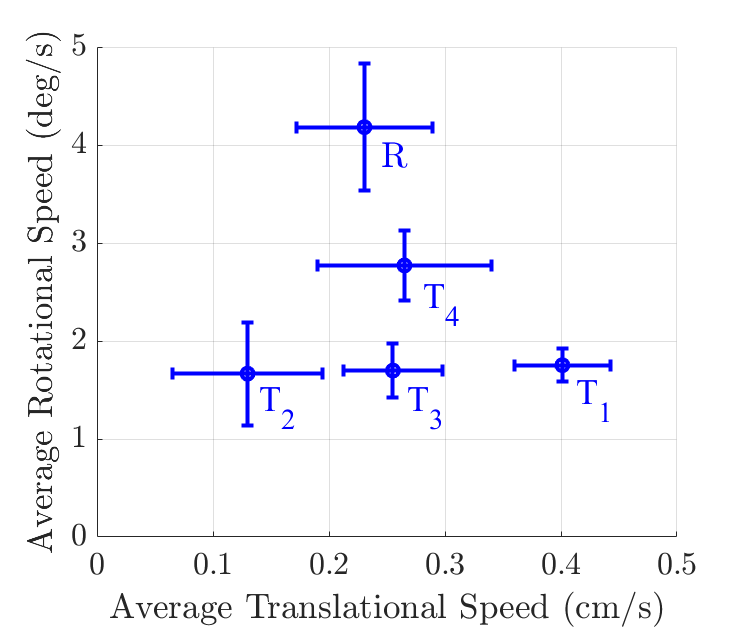} 
    \caption{(a) Gait Library comprising five gaits: rotation dominant gait $R$, and four translation dominant gaits $\{T_1,T_2,T_3,T_4\}$, where the last three gaits are permutations of the synthesized gait G. (b) Average locomotion of each gait executed for $60$ cycles. Arrows indicate the orientation of the robot after every 10 cycles. (c) The average translation and rotation speed of each gait where the bars indicate the standard deviation.}
    \label{Fig:GaitLibrary}
\end{figure*}
%%%%%%%%%%%
\subsection{Gait Library}
\label{Subsec:GaitSynthesizer}
Gaits are synthesized as described in \Sec \ref{SubSec:GaitSynthesis}. For this research, five gaits are chosen - a rotation dominant gait $R$ and a translation dominant gait $T_1$ with its permutations $T_2$, $T_3$, and $T_4$. The limb actuation patterns are shown in \Fig \ref{Fig:GaitLibrary}a. %below:
% \begin{center}
% \includegraphics[width=0.9\columnwidth]{Figures/GaitLibrarySequence.png}
% \end{center}
% The red indicate the actuated limb, while dotted black are the unactuated limbs. The numbers in bracket in front of each gait indicate the edge that is activated, e.g., `Gait G: [3 15 1]' implies that the simple cycle representing Gait G traverses edges $\{e_3,e_{15},e_1\}$ and only the $3,15,1$ elements of the gait vector $\mathbf{z}_G$ are non-zero. %
%%%%%%%%%

Each of the gaits are run for 120 gait cycles and the mean locomotion twist $\xi(\bm{p_i},\theta_i)$ is obtained that can be used by the path planner. The mean translation and rotation for each of these gaits is visually shown in \Fig \ref{Fig:GaitLibrary}. These plots highlight a key observation: the translation-dominant gaits $T_1$, $T_2$, $T_3$, and $T_4$ are offset from each other by 90 degrees as expected \textadded{(due to the limbs being positioned at 90 degree offsets)}. As the robot is fabricated to be rotationally symmetric, \textadded{we anticipated} that the twist magnitudes of these gaits would be identical; after all, they are the same gait \textadded{with actuation initiated by a different limb}. However, this is not true and highlights the sensitivity of soft robots to small manufacturing inaccuracies/non-uniformities. \textadded{While control parameters could potentially be adjusted to reduce these differences in behavior, it would be unlikely to eliminate them completely as the frictional effects appear to dominate.} Moreover, soft robots are sensitive to small changes in the environment (e.g., small bumps in the substrate) as suggested by the error bars in \Fig \ref{Fig:GaitLibrary}b. \textadded{The data-driven gait discovery and the path planning strategy accommodate these asymmetries and variations by treating each starting limb option as a distinct gait with individually (experimentally-) derived behaviors; feedback-based re-planning additionally serves to mitigate undesirable effects.}

%  \begin{figure}[h]
%     \centering
%     \begin{flushleft}
%     \textbf{\footnotesize (a)} \vspace{-10pt}
%     \end{flushleft}
%     \includegraphics[width=0.7\columnwidth]{Figures/GaitLibrary2.png}
%     \begin{flushleft}
%     \textbf{\footnotesize (b)}
%     \end{flushleft}
% \includegraphics[width=\columnwidth]{Figures/GaitCharacterization.png}    
%     \caption{Gait Library comprising five gaits : rotation dominant Gait B, and four translation dominant gaits \{G,I,J,K\}, where the last three gaits are permutations of the synthesized gait G. (a) Average locomotion of each gait executed for $60$ cycles. Arrows indicate the orientation of the robot after every 10 cycles. (b) The average translation and rotation speed of each gait where the bars indicate the standard deviation.}
%     \label{Fig:GaitLibrary}
% \end{figure}
% \begin{figure}
%     \centering    
% \includegraphics[width=0.5\columnwidth]{Figures/GaitCharacterization.png}
%     \caption{Caption}
%     \label{fig:my_label}
% \end{figure}

\subsection{Trajectory Planning}
\label{Subsec:TrajectoryPlanning}
Trajectory synthesis is computed on a binary image representation of an obstacle scenario. This input image is transcribed to a grid world cost map, $C(x, y): \mathbb{R}^2 \to \mathbb{R}$, where $\left(x, y\right)$ denote grid coordinates and $C(x, y)$ denotes the cost associated with each grid location. Obstacle locations are characterized by high costs, which fall off radially with distance. 
A tunable parameter governs how quickly cost decays and is used to effectively dilate obstacles.
Grid locations are classified as either \textit{occupied} or \textit{free} based on a pre-configured cost threshold. This results in a configuration space suitable for a point representation of the robot. Robot presence at any \textit{occupied} grid location constitutes an obstacle collision.
For a prescribed goal position $\bm{x}_{\rm goal} \in E(2)$, cost-to-go $C_{\rm go}(x, y; \bm{x}_{\rm goal}): \mathbb{R}^2 \to \mathbb{R}^+$ across the grid world is quickly computed using an expanding wavefront approach. The cost-to-go allows us to discern the relative value of potential trajectory destinations within the world.
Gait-based controlled trajectories are then designed within this grid world representation of the locomotion scenario. 

\smallskip \noindent{\bf Gait Models.}
Trajectory plans adhere to a rotate-then-translate motion paradigm. Synthesis entails searching for a sequence of these rotate-then-translate pairs, in order to move the robot from a starting pose, $g_0 \in SE(2)$, to a desired goal location, $\bm{x}_{\rm goal}$. This conceptually simplifies the trajectory planning problem; planning becomes an iterative process of: \textbf{(1)} `aiming'  (i.e. rotation), then \textbf{(2)} traveling in that direction (i.e. translation). This procedure terminates when the trajectory plan reaches  a pre-defined radius $\delta_{\rm goal}$ of the goal position  $\bm{x}_{\rm goal}$.

From the set of gaits that \MSoRo~ is able to accomplish, we select a subset to be used in trajectory planning and control. 
A single gait is selected to accomplish rotationally-dominant motion. We denote this gait as $R$; its behavior is characterized by a time-averaged body velocity twist $\xi^{\rm R} \in \mathfrak{se}(2)$, and a gait periodicity of $Q_{\rm R}$ seconds. 
The remaining $d$ gaits are characterized by translationally-dominant motion, and denoted as the set $\left\{ T_1, T_2, \ldots, T_d  \right\}$. Each translational gait $T_i$ exhibits a time-averaged body velocity twist $\xi^{\rm T_i} \in \mathfrak{se}(2)$ and gait periodicity $Q_{\rm T_i}$, where $i \in \left\{1 \ldots d \right\}$. 

\begin{figure}[t]
    \centering    
    \includegraphics[width=1.0\columnwidth]{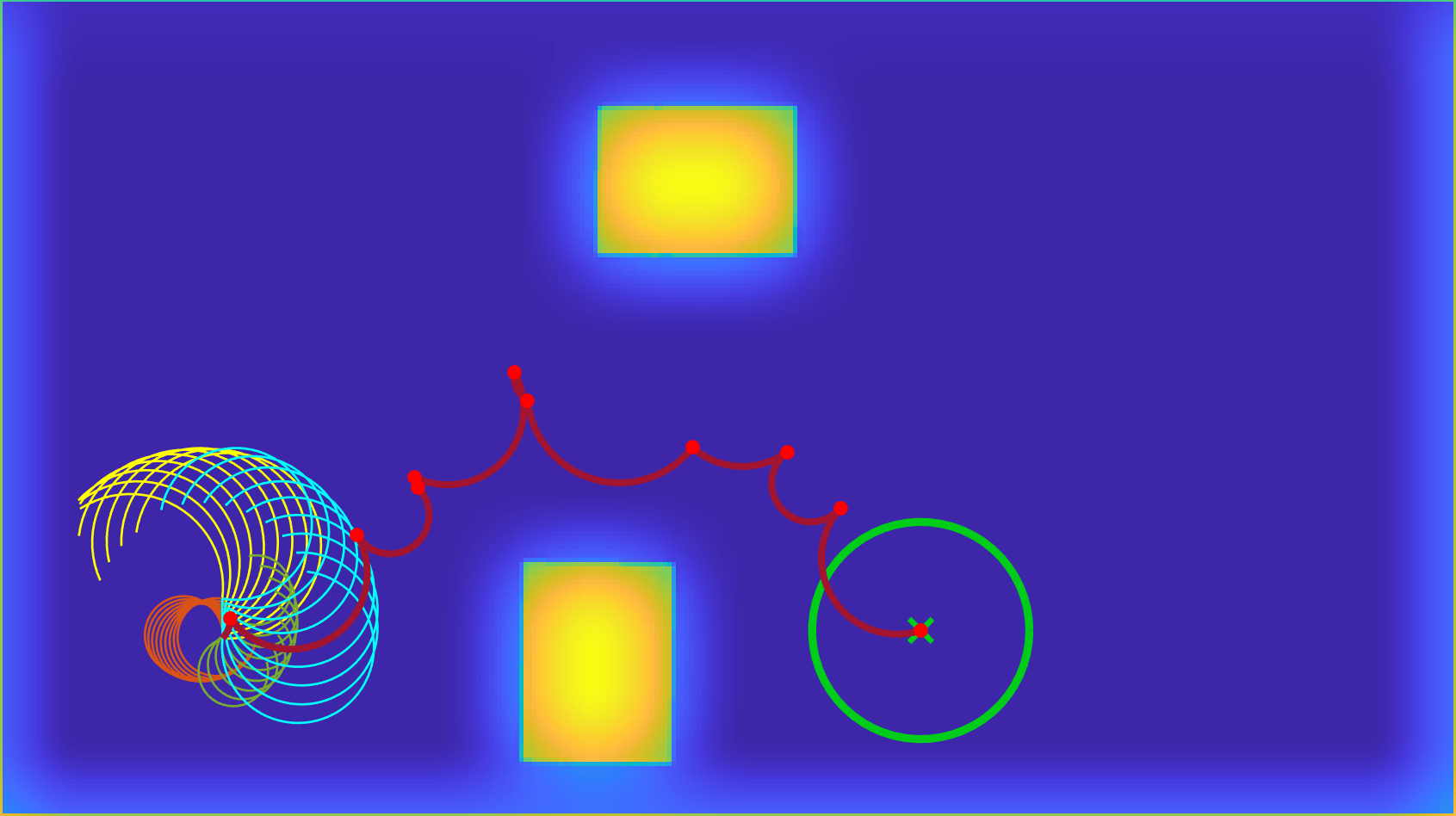}
    \caption{Controlled trajectories are computed over a grid world cost map representation of the locomotion scenario, $C(x, y)$. Obstacle locations are assigned high costs (yellow) that decay with distance; locations far from obstacles entail low cost (dark blue). The trajectory solution (dark red) moves the robot from its starting pose $g_0$ (lower-left) to a prescribed goal position $\bm{x}_{\rm goal}$ (green `X'). An example trajectory search, beginning at the starting pose, is illustrated. Cyan, yellow, light green and orange trajectories depict explored \textadded{expansions using} each translational gait, $T_1, T_2, T_3, T_4$, after initial rotation by gait R. The gait sequence leading to a subsequent pose with the least cost-to-go is selected; subsequent expansion then focuses on this new pose.}
    \label{Fig:TrajPlanner}
\end{figure}
\smallskip \noindent{\bf Trajectory Synthesis.}
The planning strategy presented here employs a greedy breadth-first search through the space of possible \MSoRo~ gait sequences. The approach successively expands sets of neighboring, collision-free trajectory destinations that may be reached by a single rotation-translation gait sequence. The neighboring destination with the smallest cost-to-go is then selected for subsequent expansion. \Fig \ref{Fig:TrajPlanner} illustrates trajectory exploration and synthesis, through an obstacle-strewn environment. This strategy produced feasible trajectories, and the corresponding gait sequences needed to accomplish them, facilitating the robot's intelligent traversal through a variety of obstacle arrangements.

Beginning at an initial expansion pose $g^{\rm expand} \in SE(2)$, the planner first samples trajectory end points that may be reached by the rotational profile $\xi^{\rm R}$, over durations of $n_{\rm R} = 1 \ldots N_{\rm R}$ rotational gait periods. $N_{\rm R} \in \mathbb{Z}^{+}$ is a fixed limit and computed such that $\xi^{\rm R}_{\omega} \cdot N_{\rm R} < \frac{\pi}{2}$. Reachable robot poses, using the rotational gait $R$, are denoted $g^{\rm R}_{n_{\rm R}} \in SE(2)$ and expressed relative to the initial expansion pose $g^{\rm expand}$. These poses are checked for collisions; if a collision is identified, the pose is discarded and removed from consideration going forward. Beginning from each collision-free $g^{\rm R}_{n_{\rm R}}$, trajectory end points are forward sampled for each translational motion profile $\xi^{\rm T_i}$ where $i = 1 \ldots d$, and durations of $n_{\rm T_i} = 1 \ldots N_{\rm T}$ translational gait periods; $N_{\rm T} \in \mathbb{Z}^{+}$ is pre-configured and denotes the maximum number of consecutive cycles a translational gait may be run. The poses of these trajectory end points, relative to $g^{\rm R}_{n_{\rm R}}$, are denoted $g^{\rm T_ i}_{n_{\rm T_i}}$. Their poses, relative to the initial expansion pose, are computed as $g^{\rm R}_{n_{\rm R}} \cdot g^{\rm T_i}_{n_{\rm T_i}}$; their corresponding spatial poses are $g{\left(n_{\rm R},i,n_{\rm T_i}\right)} = g^{\rm expand} \cdot g^{\rm R}_{n_{\rm R}} \cdot g^{\rm T_i}_{n_{\rm T_i}}$. The trajectory end points, as well as sub-sampled points along each trajectory segment, are tested for collisions; a trajectory expansion is discarded if a collision is detected. Cost-to-go $C_{\rm go}$ is evaluated at the ($x, y)$ coordinates associated to each $g{\left(n_{\rm R},i,n_{\rm T_i}\right)} \in SE(2)$. The motion sequence, described by $\{ n_{\rm R},i,n_{\rm T_i}\}$, and the corresponding trajectory destination $g{\left(n_{\rm R},i,n_{\rm T_i}\right)}$ that are associated with the lowest cost-to-go, are selected as the optimal control and subsequent node to be expanded, respectively.
This procedure iterates until the selected pose $g{\left(n_{\rm R},i,n_{\rm T_i}\right)}$ falls within a threshold distance $\delta_{\rm goal}$ of the goal position $\bm{x}_{\rm goal}$.

\subsection{Path Recalculation}
\begin{figure}[h]
\vspace*{0.065in}
    \centering
\begin{flushleft}
    {\scriptsize \textbf{(a)}} \vspace{-20pt}
\end{flushleft}    \includegraphics[width=0.9\columnwidth,align=t]{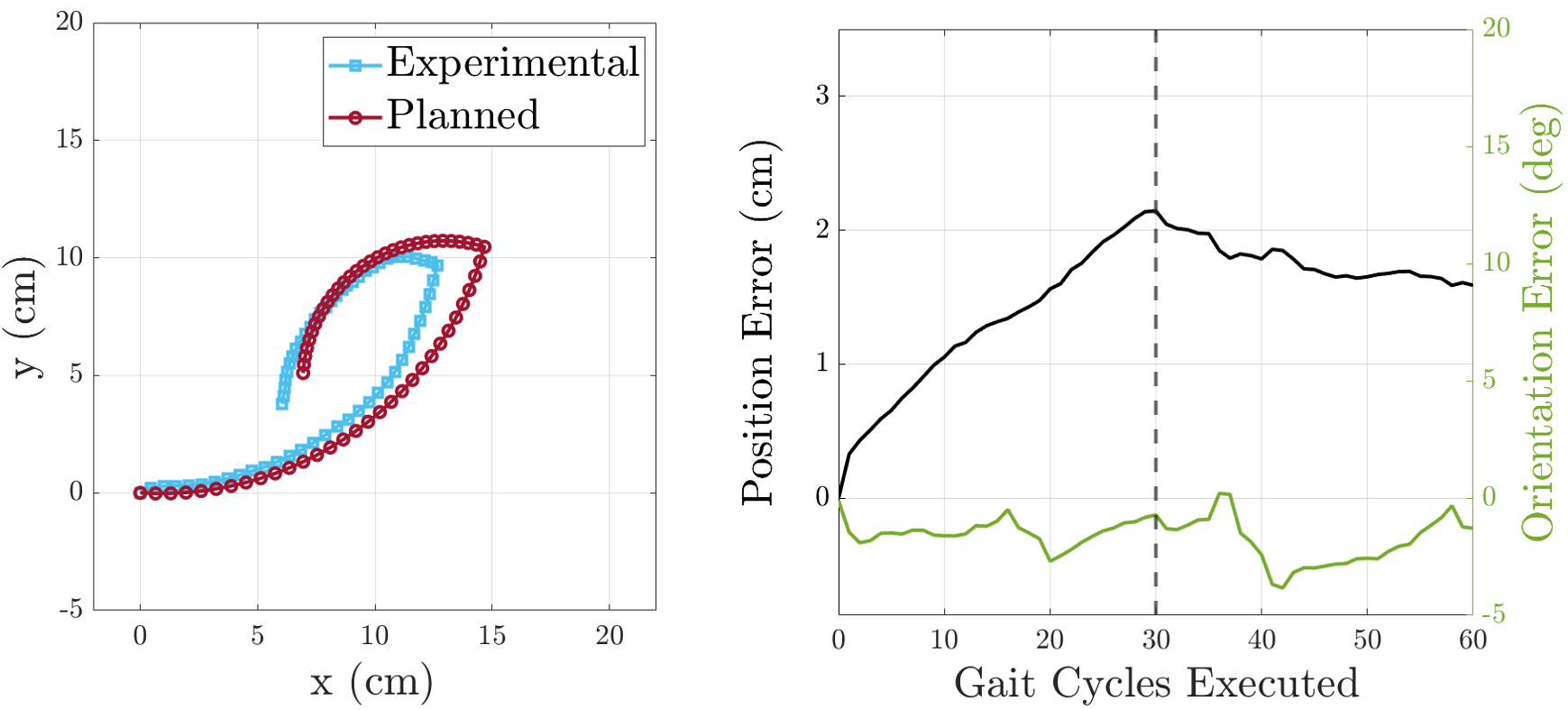}
\begin{flushleft}
    {\scriptsize \textbf{(b)}} \vspace{-20pt}
\end{flushleft}    \includegraphics[width=0.9\columnwidth,align=t]{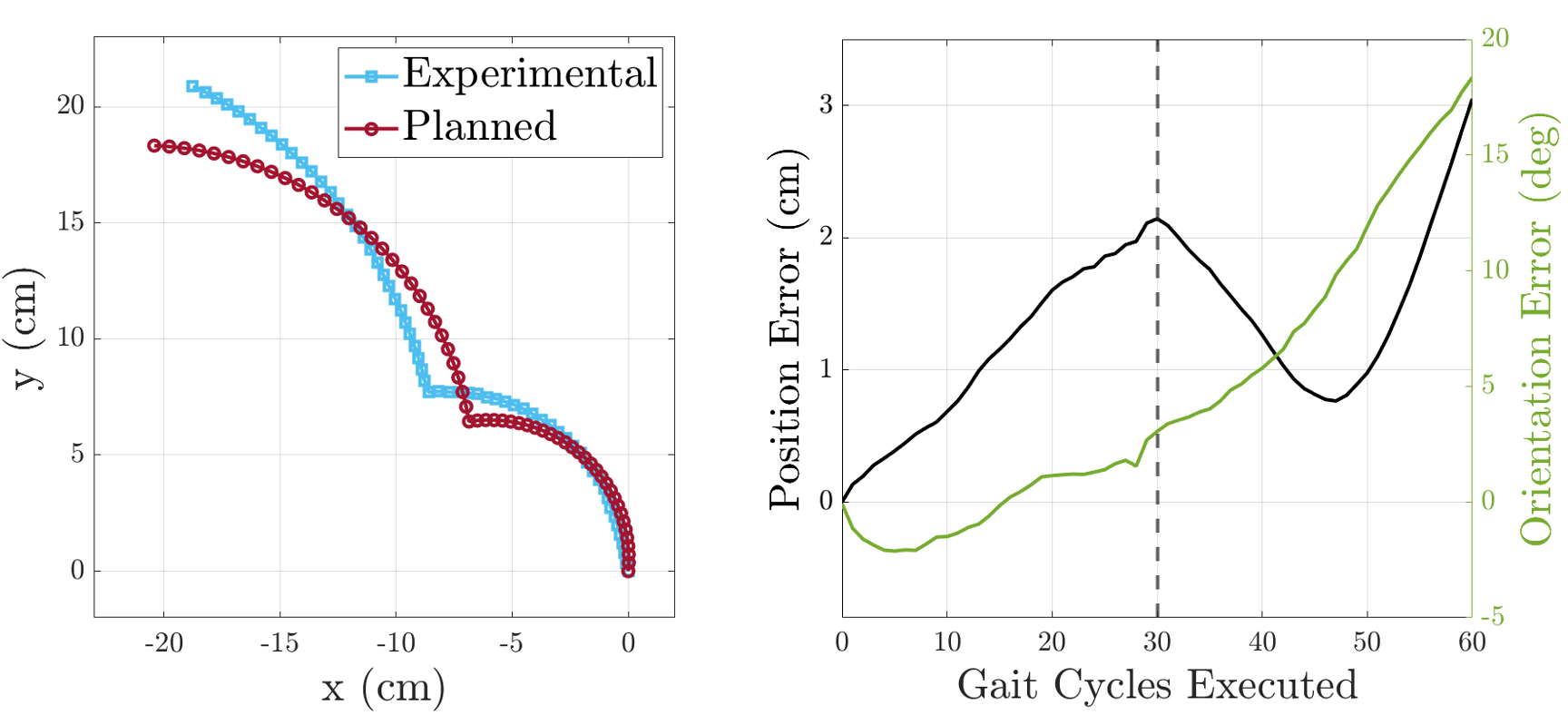}
    \caption{The error between the path-planned and experimental poses of the robot for open-loop paths consisting of 60 gait cycles. The paths include a gait switch from  (a) $T_1$ to $T_2$ and (b) $T_2$ to $T_1$ after 30 gait cycles (dashed vertical line). The plots indicate non-monotonic increase in the pose error. }
    \label{Fig:GatiSwitchingError}
\end{figure}

As observed, the locomotion gaits have both rotation and translation associated with them. As the robot performs gait cycles and switches gaits, the pose error does not always monotonically increase. Generally, the position error reaches some maximum value and then decreases. This observation is illustrated in \Fig \ref{Fig:GatiSwitchingError}. Consequently, path recalculation should be performed when the position error exceeds a user-defined error threshold upon completion of a gait sequence and/or at user-defined intervals (i.e., every $n$ gait cycles). 
%Consequently, the path-recalculation is performed upon completion of a gait sequence and then comparing the position error with the user-defined threshold error.

% \begin{figure}
%     \centering
%     \includegraphics[width=\columnwidth]{Figures/IntegrationArchitecture.png}
%     \caption{Caption}
%     \label{Fig:ParallelProgramming}
% \end{figure}
% \begin{figure}
%     \centering
%     \includegraphics[width=0.5\columnwidth]{Figures/Setup.png}
%     \caption{Caption}
%     \label{Fig:Setup}
% \end{figure}

%%%%%%%%%%%%%%%

\section{Experiments and Discussion}
\label{Sec:Experiments}
    % \subsection{Experimental Setup}
%%%%%%%%%%

%The experimental setup consists of the the \MSoRo~on a rubber garage mat, and two overhead webcams. Low-level robot control is performed by the \Arduino and is integrated with high-level planning and localization that is performed in MATLAB\textregistered. The first webcam is used to capture HD video of the experiments. The second webcam has its properties (e.g., contrast, brightness, etc.) adjusted to allow for better image segmentation of the four neon markers on the robot hub to facilitate tracking. Similarly, the obstacles, and goal are identified. The real-time robot pose is estimated as described in \Sec \ref{Subsec:Localization}. The Gait Library described in \Sec \ref{Subsec:GaitSynthesizer} informs path planning described in Sec. \ref{Subsec:TrajectoryPlanning}.

To validate the 
closed-loop path planner, we performed experiments with three world scenarios with obstacles: (1) World 1 where the robot has to perform a `zig' maneuver to go around the obstacle, (2) World 2 where the robot needs to perform a `zig' and `zag' motion to avoid obstacles and reach the goal, and (3) World 3 where the robot needs to go between two obstacles to reach the goal. All experiments are performed with the experimental setup and methodology discussed in \Sec \ref{Sec:Methodology}. A rubber garage mat is used as the substrate and paper cutouts are used for the obstacles. The \MSoRo~successfully maneuvers past the obstacles to reach the target position in all three scenarios. \textadded{The results discussed in this section are visualized in the multimedia attachment to this paper.}%The results are discussed \textadded{and also captured in the multimedia attachment to this paper.}\textadded{\st{, and the reader can also refer to the multimedia attachment to the paper for the videos.}}

\begin{figure*}[t]
\vspace*{0.09in}
% trim= {left bottom right top}
    % \includegraphics[width=0.35\textwidth,trim={0pt 0pt 500pt 0pt},clip,align=t]{Figures/Zag-Setup.png} 
    % \includegraphics[width=0.3\textwidth,trim={750pt 0pt 750pt 0pt},clip,align=t]{Figures/Zag-Recalculation.png} 
    % \includegraphics[width=0.3\textwidth,trim={250pt 0pt 250pt 0pt},clip,align=t]{Figures/Zag-Error.png}
\textbf{\scriptsize(a)}\vspace{-10pt}\\
\begin{minipage}
{0.3\textwidth}
\includegraphics[width=\columnwidth,trim={0pt 0pt 00pt 0pt},clip,align=t]{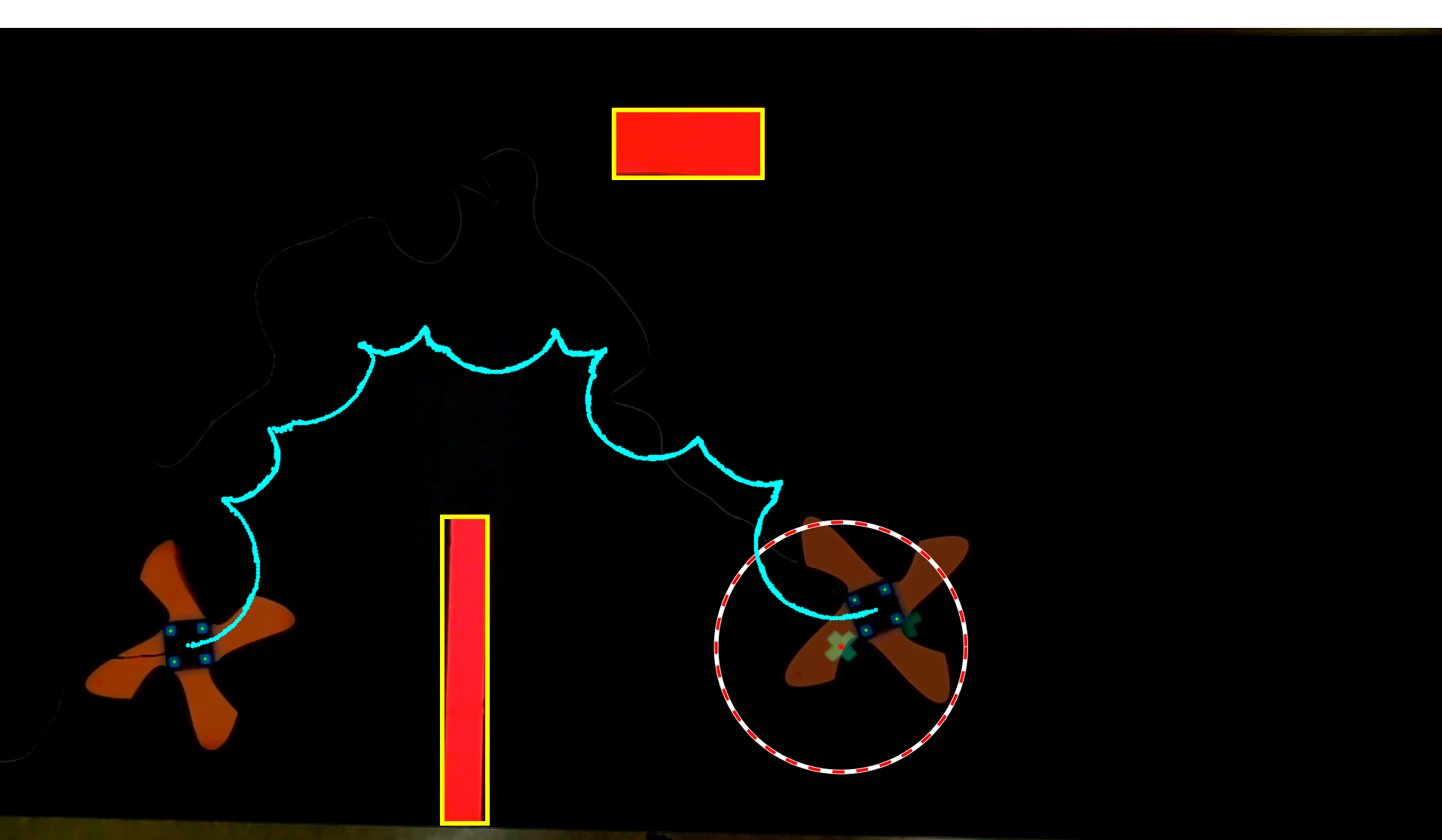} 
\end{minipage}
\hspace*{10pt}
\begin{minipage}{0.32\textwidth}
\centering
\includegraphics[width=0.8\columnwidth,clip,align=t]{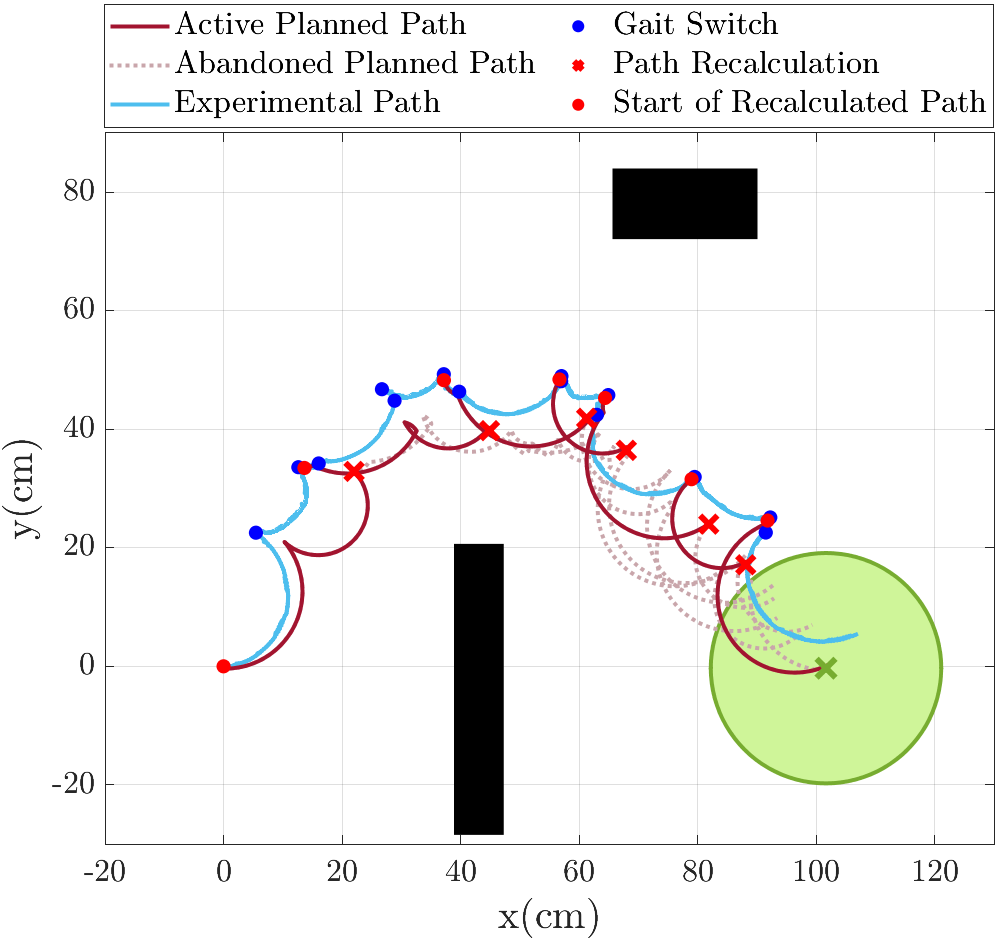}
\end{minipage}
\hspace*{10pt}
\begin{minipage}{0.28\textwidth}
\includegraphics[width=\columnwidth,clip,align=t]{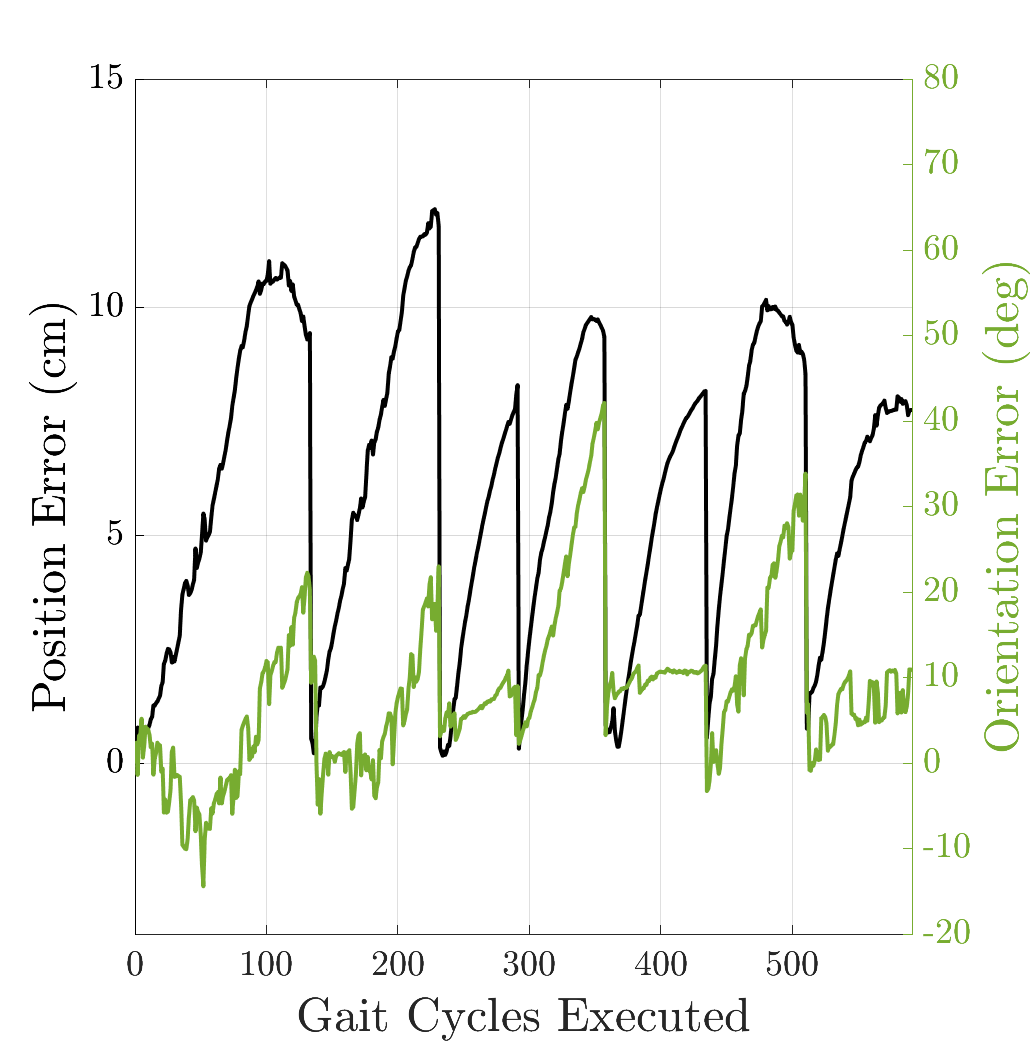}
\end{minipage}

\vspace*{10pt}
\textbf{\scriptsize(b)}\\
\begin{minipage}{0.3\textwidth}
% \textbf{\scriptsize(b)}
\includegraphics[width=\columnwidth,trim={0pt 0pt 100pt 0pt},clip,align=t]{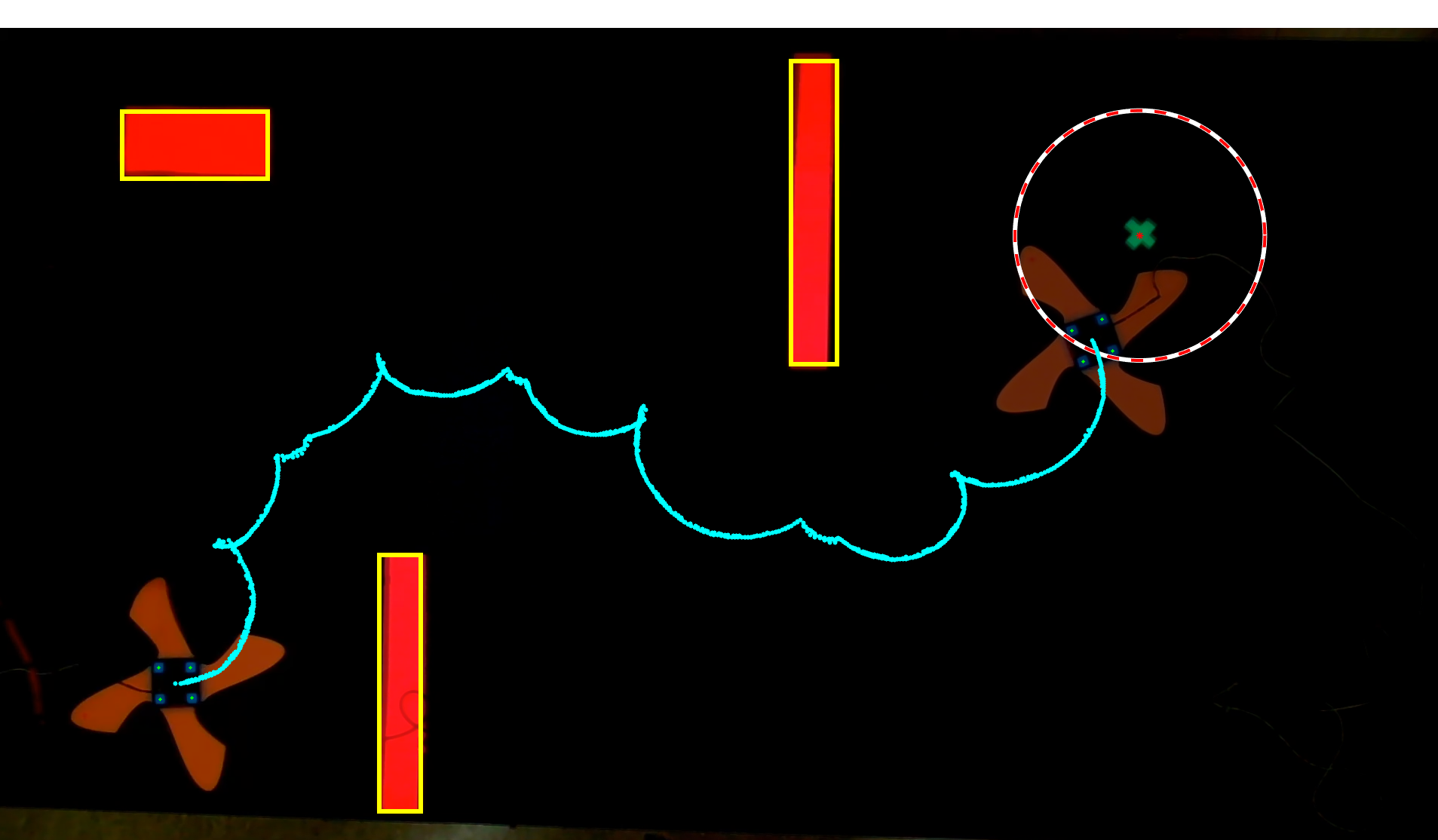} 
\end{minipage}
\hspace*{10pt}
\begin{minipage}{0.32\textwidth}
\includegraphics[width=\columnwidth,clip,align=t]{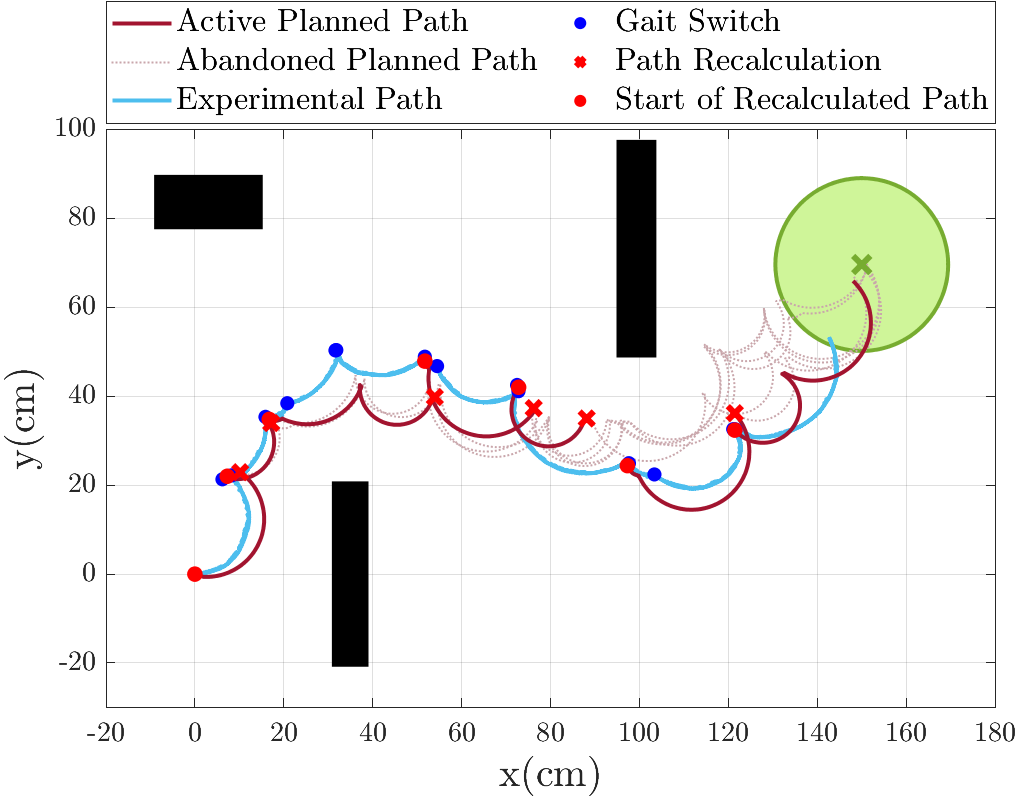}
\end{minipage}
\hspace*{10pt}
\begin{minipage}{0.28\textwidth}
\includegraphics[width=\columnwidth,clip,align=t]{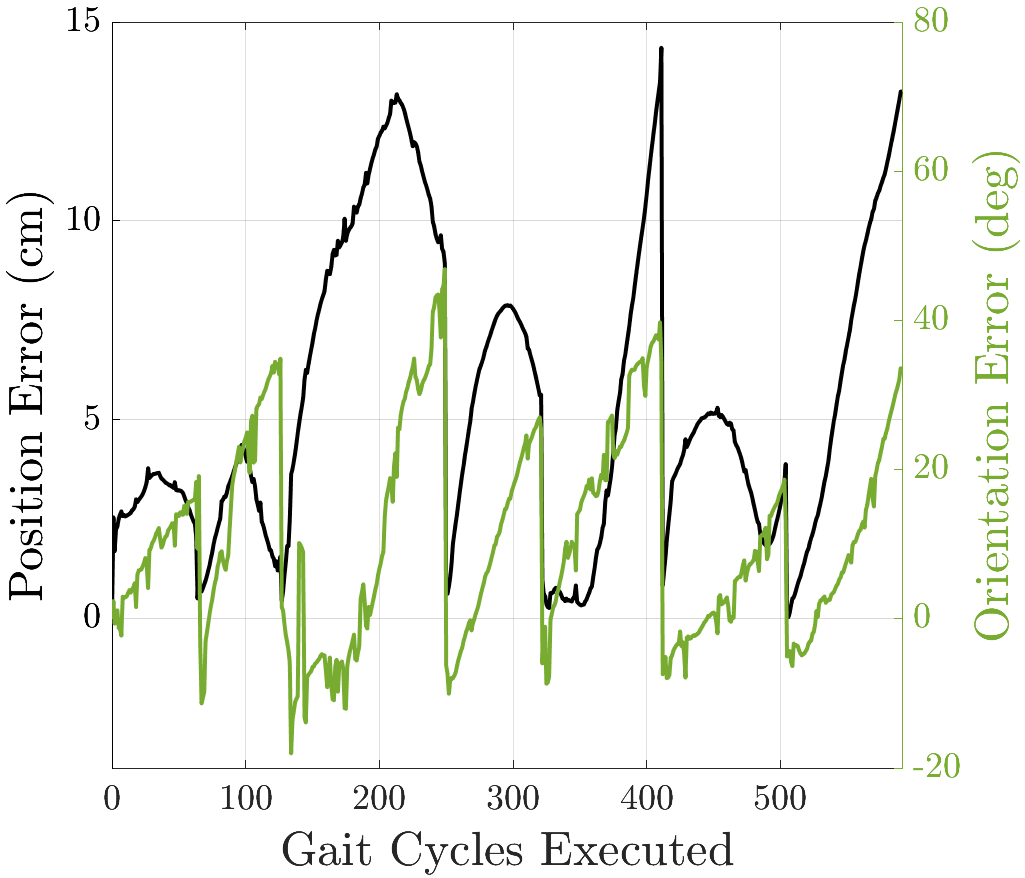}
\end{minipage}

\vspace*{10pt}
\textbf{\scriptsize(c)}\\
\begin{minipage}{0.3\textwidth}
% \textbf{\scriptsize(b)}
\includegraphics[width=\columnwidth,trim={0pt 0pt 0pt 0pt},clip,align=c]{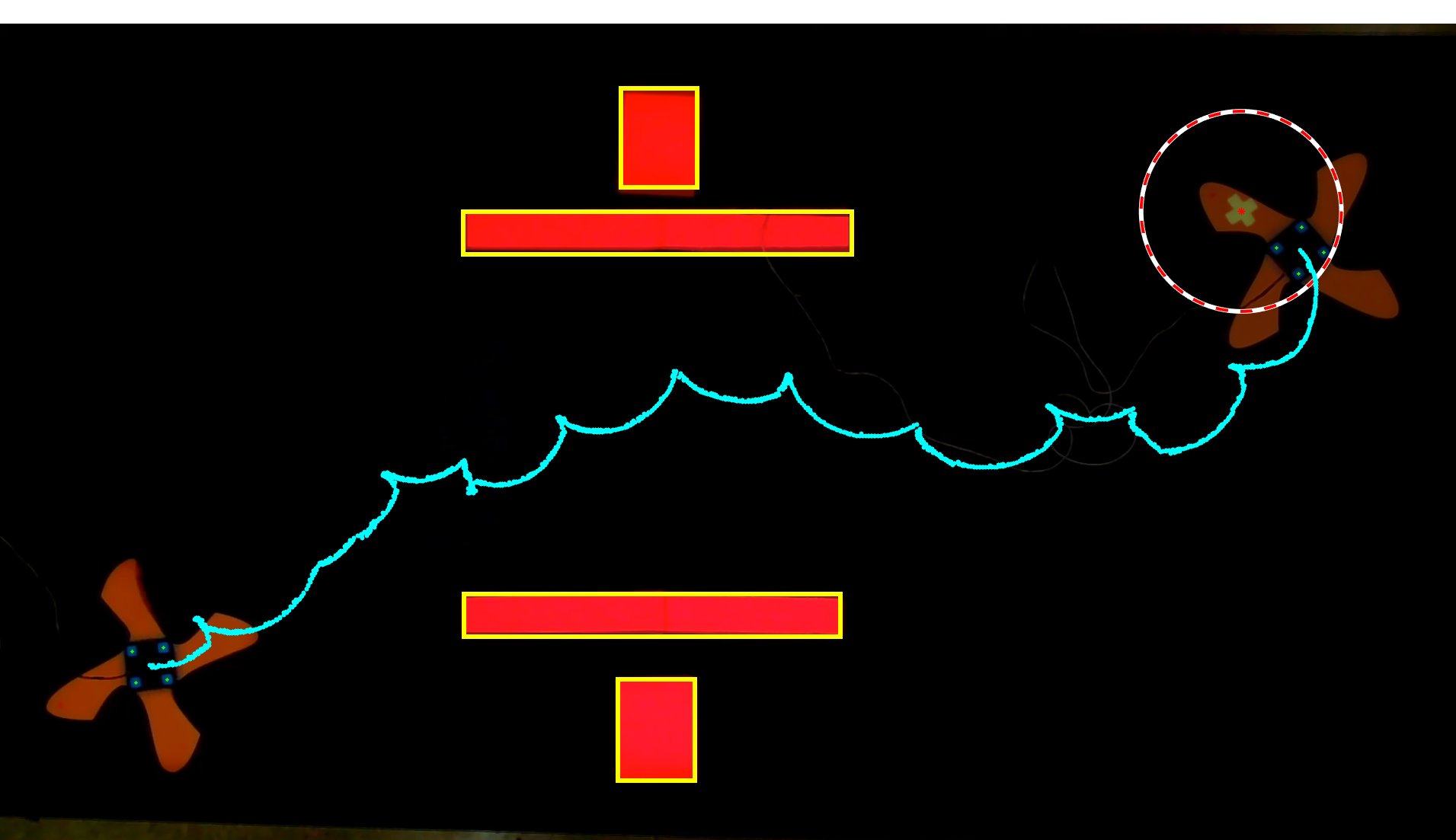} 
\end{minipage}
\hspace*{10pt}
\begin{minipage}{0.32\textwidth}
\includegraphics[width=\columnwidth,clip,trim={40pt 50pt 75pt 0pt},align=c]{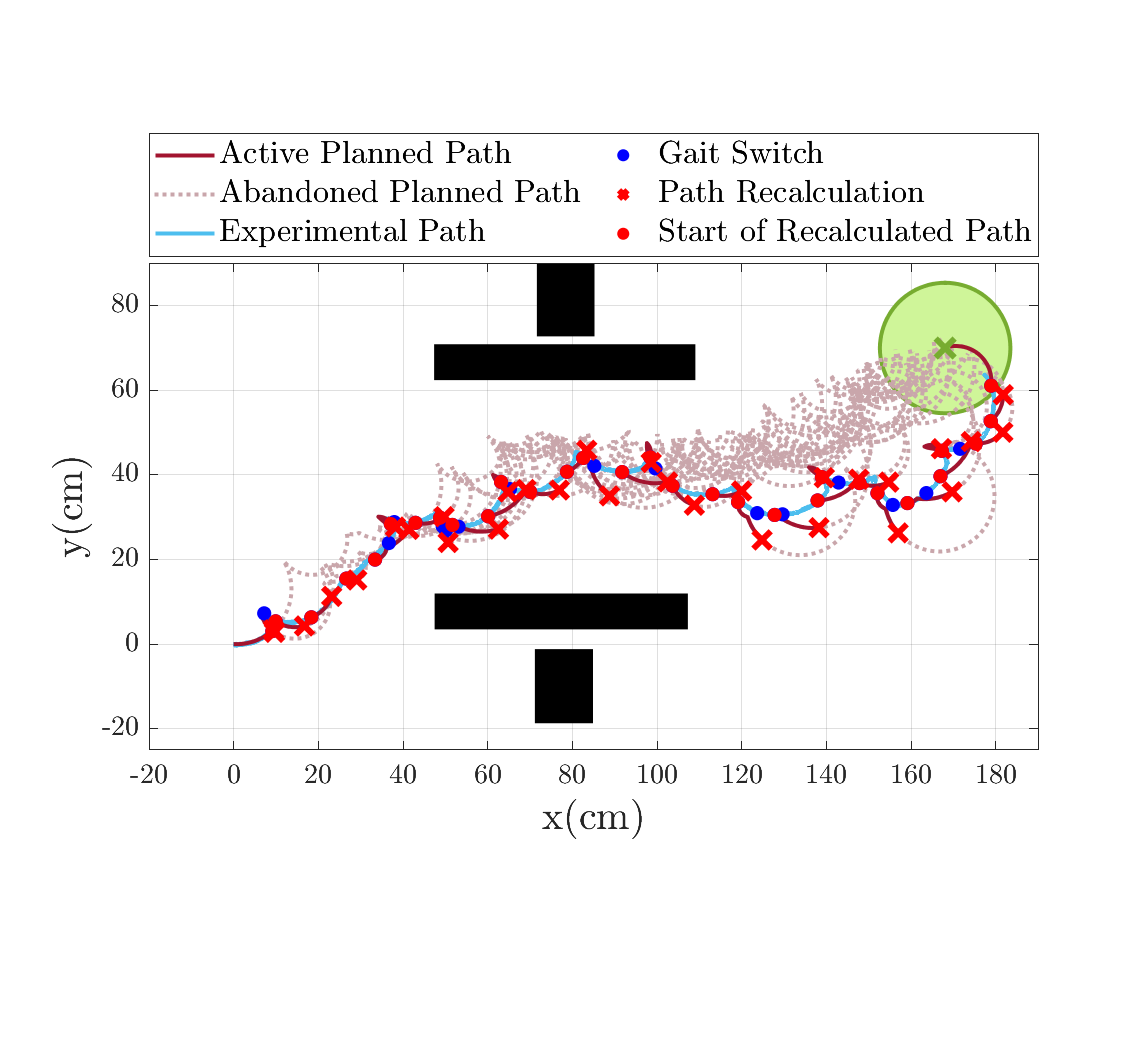}
\end{minipage}
\hspace*{10pt}
\begin{minipage}{0.28\textwidth}
\includegraphics[width=\columnwidth,clip,align=c]{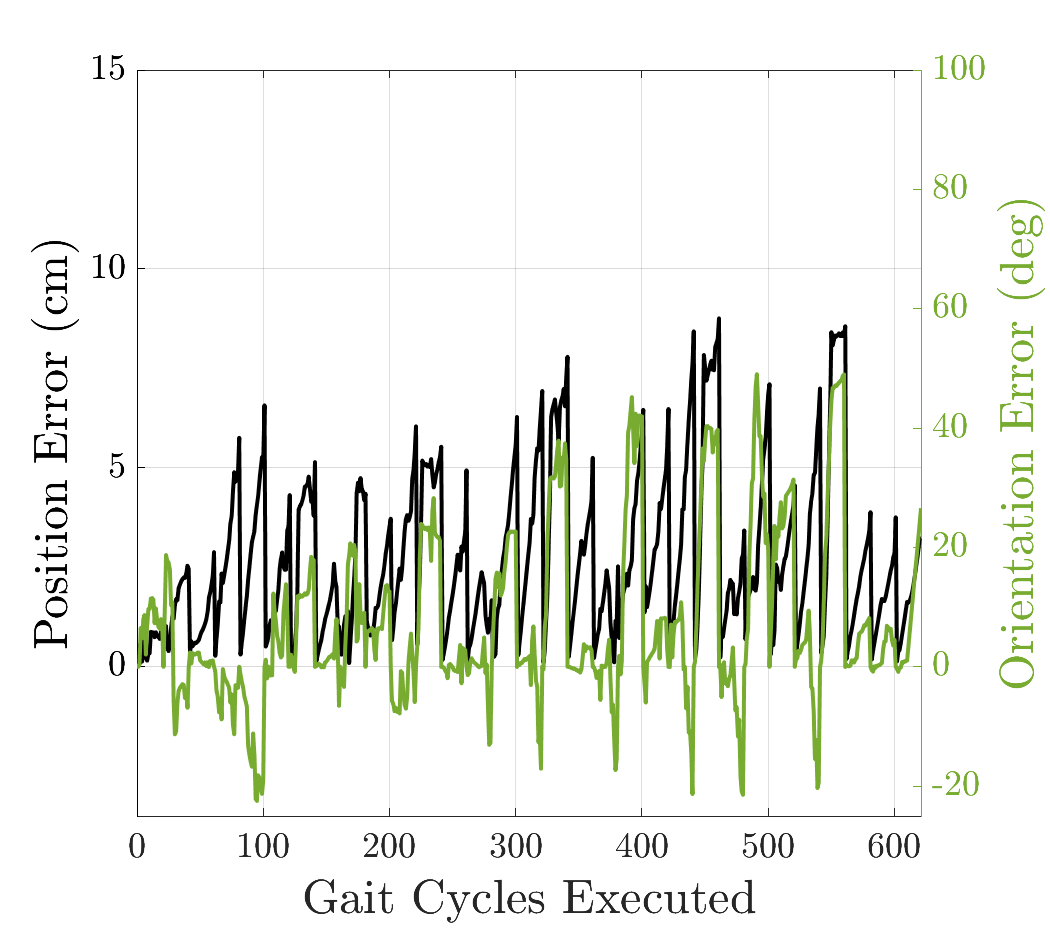}
\end{minipage}
\vspace*{-10pt}
    \caption{Example implementations of the closed loop path planner for (a) World 1, (b) World 2, and (c) World 3. The images on the left show the starting and ending robot poses and its path (cyan) that avoids the obstacles (red) to reach the acceptance boundary (circle) around the target (green `X'). The plots in the middle show the same experimental path of the robot (blue) compared to the planned path (maroon). The target is marked as a green `X' with a green acceptance circle boundary around it. This plot also marks the moments of gait switches (blue dots), path recalculations (red), and planned paths (dotted pink) that are abandoned when re-planning occurs. The plots on the right show the position and orientation errors for these paths. Sudden corrections in the error indicate re-planning.} %\MSoRo~is shown in orange and the obstacles in red are highlighted by yellow outlines. The goal is shown as the green cross and the success boundary is shown as a circle around it. The planned path and the traveled paths are shown in maroon and cyan, respectively. Each of the selected time slices indicate the instance before and after path re-planning. \hl{Fix visualization, simplify for impact}}
    \label{Fig:ClosedLoop1}
\end{figure*}

World 1 consists of two obstacles where the robot is required to maneuver one obstacle to reach the target. The planner recalculates the path six times before it reaches the goal, as shown in \Fig \ref{Fig:ClosedLoop1}a. The path is recalculated after the execution of a gait sequence if the position error exceeds a threshold. From the error plot, sudden drops in the error indicate path re-planning.
%reaches the target/goal. 
% In world 3 the  robot has to traverse through a much narrow path in between the obstacles 
%
World 2 requires the robot to perform two obstacle-avoiding maneuvers to reach the goal as shown in \Fig \ref{Fig:ClosedLoop1}b. World 2 also triggers six path recalculations. In both World 1 and World 2, the position error both increases and decreases during the execution of gait sequences; it does not consistently monotonically increase before suddenly dropping at recalculation points. Thus, the robot is allowed to complete the current gait sequence, until the point of a gait switch. If the position error still exceeds the threshold, re-planning occurs.

The World 3 scenario explores the ability of \MSoRo~to navigate narrow spaces, \Fig \ref{Fig:ClosedLoop1}c. Here, to perform this delicate maneuver, the path re-planning is triggered after every 20 gait cycles. As observed, the position errors are smaller relative to previous world scenarios. However, this scenario also involves thirty recalculations and does not appear to exploit the potential decreases in the error that were seen in the previous worlds.%

\section{Conclusion and Future Work}
\label{Sec:Conclusion}
% The research proposes a generic, closed-loop control framework for terrestrial soft robots. The experimental setup involves system integration of high-level (online path planning and offline gait synthesizer) control, low-level control (\MSoRo~actuation), and real-time pose estimation that uses parallel architecture to process visual feedback. To the best of our knowledge, this is the first instance of such system integration and experimental validation of closed loop path planning for motor-tendon actuated soft robots. {The path planning is performed using a lattice-based algorithm using the robot gaits as the inputs. The greedy breadth-first strategy generate trajectories comprise of rotate-then-translate gait pairs to iteratively solve the problem. The path is re-planned when the position error between the estimated and the executed path are above a pre-defined threshold.

For the first time, this research presents successful closed-loop path planning and obstacle avoidance of the four-limb motor-tendon actuated soft robot \MSoRo. The experiment uses real-time visual feedback from low-cost webcams to perform localization, while the lattice-based path planner generates collision-free trajectories using a greedy breadth-first approach. As soft robots are more sensitive to factors like  changes in the environment and manufacturing uncertainties, the locomotion gaits are synthesized using a data-driven environment-centric framework. Conceptually, this approach discretizes the factors dominating the environment-robot interaction and synthesizes the tracked motion of these interactions to find translation and rotation gaits. The path planner generates sequences of gaits that are pairs of rotation-then-translation. The synthesized gaits for the \MSoRo~have coupled translation and rotation. Consequently, the path is recalculated when the position error exceeds a threshold after completion of a gait sequence, or at a user-defined interval. The framework is validated on complex world scenarios with obstacles that require the robot to perform challenging maneuvers. The non-uniform nature of the surface/environment and the intentional slipping action of the robot limbs further complicate this challenge.

Future work of this research involves extending the framework to a larger gait library that allows for locomotion on different surfaces. Furthermore, work will be done to adapt the path planner to incorporate the probabilistic nature of locomotion gaits.

% \begin{itemize}
%     % \item this was a successful closed-loop, low-cost, generic planner.
%     % \item Re-planning cannot be purely based on translation error because of the coupled rotation/translation . 
%     % \item Because soft robots are more sensitive to changes in the environment, manufacturing defects, it is important to have a data-driven 
%     % \item sources of variation in the gait (tether, surface defects, actuation variance, nonlinear properties, not considering movements of transitions between gaits)
%     % \item robustness of method has been tested on a particularly difficult example: coupled rotation/translation, slipping, complicated geometry/friction
%     \item future works: adding probability into the planner
% \end{itemize}
% \balance 
\bibliographystyle{IEEEtran}  \balance
\bibliography{IEEEabrv,refs}

\end{document}